\documentclass[runningheads]{llncs}

\usepackage{eccvabbrv}

\usepackage{graphicx}
\usepackage{booktabs}

\usepackage[accsupp]{axessibility}  % Improves PDF readability for those with disabilities.

\usepackage{hyperref}

\usepackage{orcidlink}
\usepackage{times}
\usepackage{latexsym}
\usepackage{booktabs}
\usepackage{multirow} 
\usepackage[T1]{fontenc}
\usepackage[utf8]{inputenc}

\usepackage{microtype}

\usepackage{inconsolata}

\usepackage{graphicx}
\usepackage{amsmath}
\usepackage{amssymb}
\usepackage{algorithm}
\usepackage{algorithmic}
\usepackage{enumitem} 
\usepackage{comment}

\usepackage{booktabs}      % For \toprule, \midrule, \bottomrule
\usepackage{multirow}      % For multi-row cells (e.g., rotated labels)
\usepackage{multicol}      % For \multicolumn
\usepackage{subcaption}
\usepackage{graphicx}      % For \resizebox and general graphics
\usepackage{xcolor}        % For colored text (\textcolor)
\usepackage{colortbl}      % For colored table cells if needed
\usepackage{array}         % For better column alignment and spacing
\usepackage{makecell}      % For improved cell formatting
\usepackage{rotating}      % For \rotatebox
\usepackage{caption}       % For caption formatting
\usepackage{setspace}      % (optional) Adjust line spacing if caption too tight

\usepackage{fontawesome5} 
\usepackage[most]{tcolorbox}
\definecolor{BetterGreen}{HTML}{1a7f37}
\definecolor{BetterBlue}{HTML}{1f75cb}
\definecolor{BetterYellow}{HTML}{f5a800}
\definecolor{BetterRed}{HTML}{e00000}

\usepackage{wrapfig}
\usepackage[numbers]{natbib}
\definecolor{promptgray}{RGB}{235,235,235}

\begin{document}

% ---------------------------------------------------------------
% TODO REVIEW: Replace with your title
\title{Wan-R1: Verifiable-Reinforcement Learning for
Video Reasoning} 

% TODO REVIEW: If the paper title is too long for the running head, you can set
% an abbreviated paper title here. If not, comment out.
\titlerunning{Abbreviated paper title}

\author{Ming Liu\inst{1} \and
Yunbei Zhang\inst{2} \and
Shilong Liu\inst{3} \and
Liwen Wang\inst{1} \and
Wensheng Zhang\inst{1}}
\authorrunning{M.~Liu et al.}
\institute{Iowa State University, Ames, IA 50011, USA \and
Tulane University, New Orleans, LA 70118, USA \and
Princeton University, Princeton, NJ 08544, USA}
\maketitle

\begin{abstract}
Video generation models produce visually coherent content but struggle 
with tasks requiring spatial reasoning and multi-step planning. 
Reinforcement learning (RL) offers a path to improve generalization, 
but its effectiveness in video reasoning hinges on reward design---a 
challenge that has received little systematic study. We investigate 
this problem by adapting Group Relative Policy Optimization (GRPO) to 
flow-based video models and training them on maze-solving and robotic 
navigation tasks. We first show that multimodal reward models fail 
catastrophically in this setting. To 
address this, we design \emph{verifiable reward functions} grounded in 
objective task metrics. For structured game environments, we introduce 
a multi-component trajectory reward. For robotic navigation, we 
propose an embedding-level verifiable reward. Our experiments show that RL 
fine-tuning with verifiable rewards improves generalization. For example, on complex 
3D mazes, our model improves exact match accuracy by 29.1\% over the SFT baseline, and on trap-avoidance tasks by 51.4\%. Our systematic 
reward analysis reveals that verifiable rewards 
are critical for stable training, while multimodal reward models could lead to 
degenerate solutions. These findings establish verifiable reward design 
as a key enabler for robust video reasoning. Code will be public available.
\end{abstract}

% ============================================
% 1. INTRODUCTION
% ============================================
\section{Introduction}

The emergence of video generation models, such as Veo~\citep{GoogleVeo2ModelPage} and Sora~\citep{openai2025sora2}, offers a new approach to visual reasoning. Instead of generating chains of thought through text tokens~\citep{GPT5ChatGPT}, these models reason by generating temporally coherent visual sequences~\citep{wiedemer2025videomodelszeroshotlearners}. This ``reasoning via video'' paradigm naturally captures spatial relationships, physical dynamics, and temporal causality within a continuous visual medium.

Recent work~\citep{wiedemer2025videomodelszeroshotlearners,guo2025videomodelsreadyzeroshot} shows that video models can learn reasoning tasks like maze-solving. Another work studys whether the same \emph{reasoning-via-video} paradigm can support \emph{real-world} semantic navigation, by evaluating on Target-Bench~\citep{wang2025targetbench}. These tasks require spatial perception, path planning, and multi-step decision making, which remain challenging for current models~\citep{guo2025video}. Supervised fine-tuning (SFT) on demonstration videos teaches models to solve specific mazes, but it exposes a critical limitation: SFT models generalize poorly to out-of-distribution scenarios~\citep{yang2025reasoningvideoevaluationvideo}. A model trained on easy mazes struggles with harder configurations. Training on regular grids does not transfer to irregular or 3D layouts, and changes in visual texture often break performance. This brittleness suggests that SFT encourages the memorization of training patterns rather than robust, transferable reasoning.

We investigate whether reinforcement learning (RL) can bridge this generalization gap. Because RL explores diverse solution strategies during training, it can encourage models to learn generalizable reasoning skills rather than surface-level heuristics. We adapt flow Group Relative Policy Optimization (GRPO)~\citep{liu2025flow}, a recent advance in RL for generative models, to flow-based video generation.

\begin{figure*}[h]
    \centering
    \vspace{-1.5em}
    \includegraphics
    [width=0.85\textwidth]{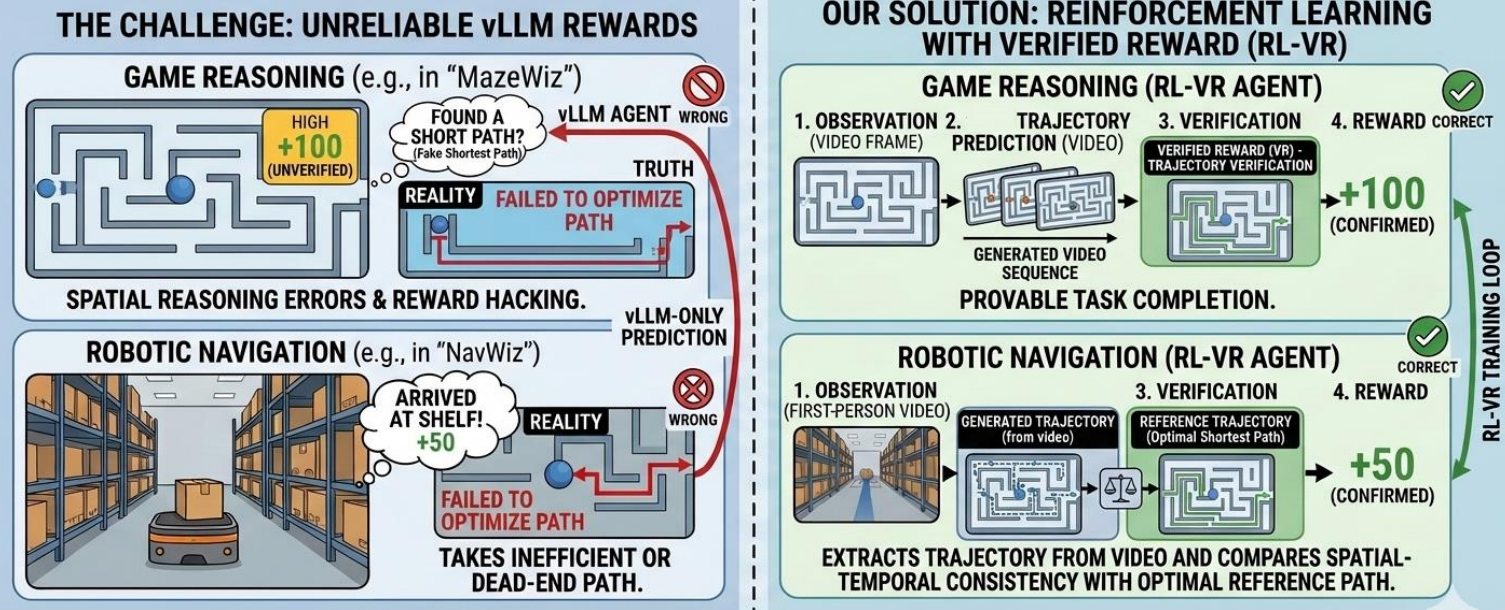}
    \vspace{-1em}
    \caption{VLLM fails to provide reliable reward, verified rewards can guarantee correct supervision.}
    \vspace{-2em}
    \label{fig:method}
\end{figure*}

Applying RL to video reasoning requires careful reward design. Multimodal reward models are susceptible to reward hacking~\citep{zhou2025generativerlhfvlearningprinciples,liu2025is,hong2026understandingrewardhackingtexttoimage}; the generator learns to produce visually convincing videos that satisfy the judge without actually solving the underlying task. We avoid this by using verifiable reward functions. For game environments, we extract the agent's path from the generated video and compare it against the ground-truth solution, ensuring the reward reflects true performance without the pitfalls of learned evaluators. For robotic navigation, where ground truth takes the form of reference rollout videos rather than discrete action sequences, we design an embedding-level reward that measures frame-wise visual similarity, temporal consistency, and endpoint fidelity between generated and reference videos. Our contributions are as follows:
\begin{itemize}
\item We adapt Flow-GRPO to flow-matching image-to-video generation 
with denoising-step reduction, making online RL for video reasoning
computationally practical.

\item We provide a systematic study of reward design for video 
generation RL. We show that vision-language reward models fail via 
reward hacking and that sparse task-completion rewards are insufficient. 
We introduce multi-component verifiable rewards based on trajectory 
extraction.

\item We propose an embedding-level verifiable reward for robotic video 
reasoning that compares generated rollouts against reference 
demonstrations via frame similarity, temporal-order consistency, and 
endpoint fidelity. This reward enables RL training for real-world 
navigation without discrete action supervision.

\item We demonstrate that RL with verifiable rewards consistently 
outperforms SFT baselines, improving exact match accuracy by up to 51.4 percentage 
points on out-of-distribution maze types. On Target-Bench for robotic 
path planning, our model achieves the best navigation score, reducing 
displacement errors by roughly half.

%%NEW: Added analysis contribution
\item We provide practical insights for scaling RL in video reasoning: 
we analyze KL regularization, denoising reduction, cold-start 
requirements, and test-time scaling, offering guidelines for 
practitioners applying RL to video reasoning.
\end{itemize}

% ============================================
% 2. RELATED WORK
% ============================================
\section{Related Work}

\subsection{Video Generation Models}

The field of video generation has seen rapid progress. Sora-2 \citep{brooks2024video} demonstrated the ability to produce controllable, physically plausible videos with coherent audio and dialogue integration. Closed source models—including Runway's Gen-3 \citep{runway2024gen3}, Pika Labs \citep{pikalabs2024pika15}, Luma AI \citep{lumalabs2024dreammachine}, and Google DeepMind's Veo family \citep{deepmind2024veo2, deepmind2025veo3}—have pushed visual fidelity and realism further, though their implementations are not publicly available. Meanwhile, open-source alternatives like Stable Video Diffusion \citep{blattmann2023stable}, OpenSora \citep{opensora}, Hunyuan-Video \citep{kong2024hunyuanvideo}, and the Wan series \citep{wan2025wan} have broadened community access by providing efficient model designs and scalable training pipelines for high-quality video generation.

\subsection{Reasoning via Visual Generation}

Chain-of-Thought (CoT) prompting has substantially improved the reasoning capabilities of language models~\citep{wei2022chain, wang2022self, guo2025deepseek}. By incorporating reinforcement learning, CoT-style reasoning can be directly embedded into the training process, allowing models to develop internalized multi-step reasoning abilities~\citep{openai2025o3systemcard}. Recently, the emergence of unified architectures capable of both generation and comprehension has introduced a novel reasoning framework built around \emph{interleaved vision-language outputs}~\citep{xie2025show, wu2025qwen, duan2025got}, offering a more visually grounded and expressive approach to tackling complex multimodal reasoning tasks. Recent studies also explore the potential of video
generative models in video reasoning~\citep{guo2025videomodelsreadyzeroshot,wiedemer2025videomodelszeroshotlearners,yang2025reasoningvideoevaluationvideo,wang2025targetbench}.

\subsection{RL for Generative Models}
Reinforcement learning has demonstrated remarkable effectiveness in aligning large language models (LLMs) with human preferences \citep{ouyang2022training,christiano2017deep}. Group Relative Policy Optimization (GRPO) \citep{guo2025deepseek} has recently emerged as a scalable alternative to PPO~\citep{schulman2017proximalpolicyoptimizationalgorithms} for reasoning tasks. In the domain of visual generation, PPO-style policy gradients also demonstrated impressive generalization~\citep{black2024trainingdiffusionmodelsreinforcement,fan2023dpokreinforcementlearningfinetuning,gupta2025simpleeffectivereinforcementlearning}. More recently, DanceGRPO \citep{xue2025dancegrpo} and Flow-GRPO \citep{liu2025flow} pioneered the adaptation of GRPO to visual synthesis, providing a unified framework for diffusion and flow models through SDE reformulation and achieving stable training across text-to-image, text-to-video, and image-to-video generation.

% \subsection{World Models for Robotic Planning}

% World models have long been proposed as a substrate for planning and control, but evaluating whether a video generator is \emph{useful for planning} requires geometry-aware metrics beyond perceptual quality. Target-Bench~\citep{wang2025targetbench} introduces a real-world benchmark for mapless path planning toward text-specified semantic targets: a model predicts a future egocentric video, a world decoder recovers camera motion, and the resulting path is compared against SLAM-verified ground truth. In our Target-Bench setting, we follow this protocol and use VGGT~\citep{wang2025vggt} to decode camera trajectories from generated videos.

% ============================================
% 3. METHOD
% ============================================
\section{Method}

We present our approach for improving video generation reasoning through reinforcement learning. We first describe the task formulation, then detail our adaptation of GRPO to flow-based video models, and finally present our verifiable reward design.

\begin{figure*}[h]
    \centering
    % \vspace{-1em}
    \includegraphics
    [width=0.85\textwidth]{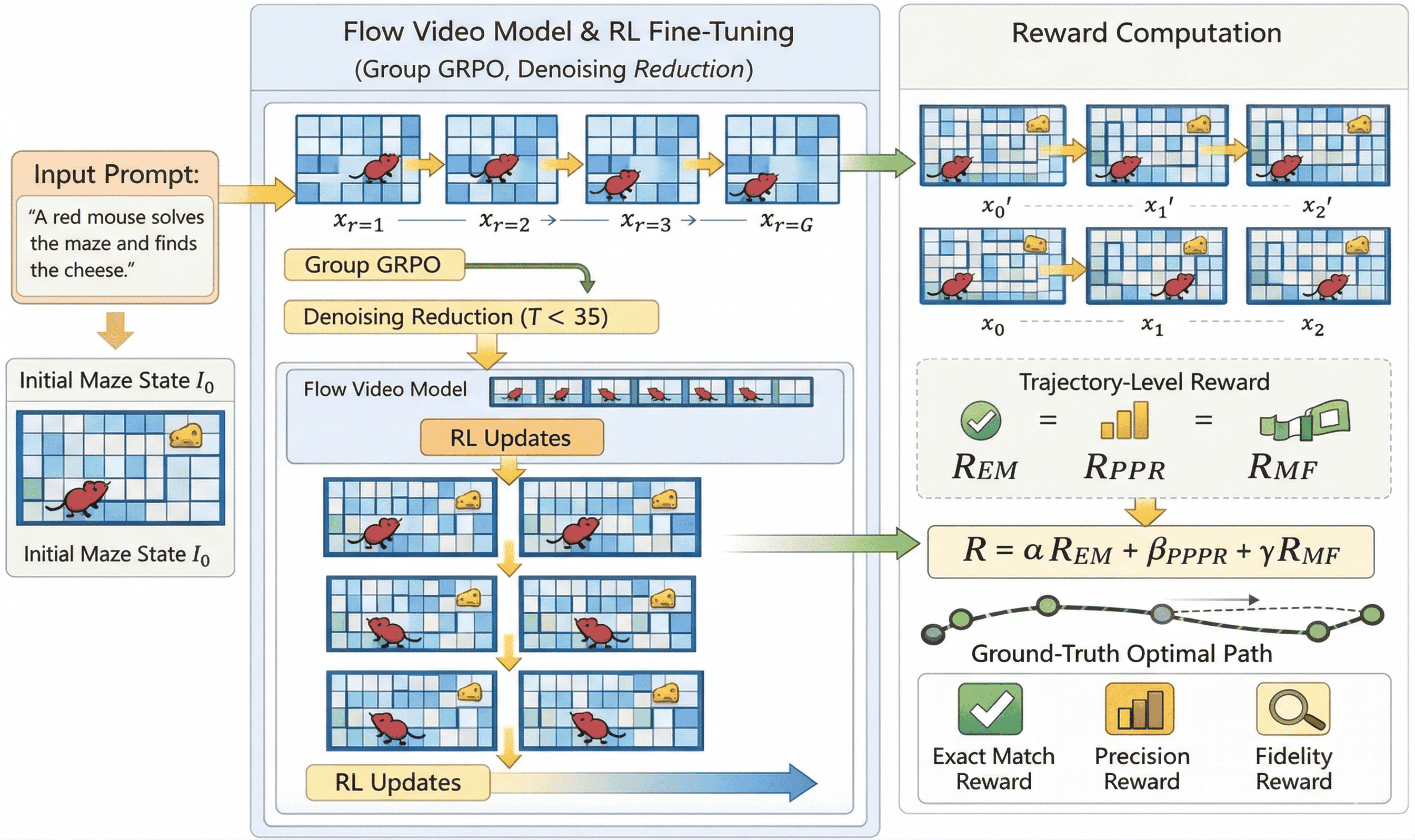}
    \vspace{-1em}
    \caption{Flow-GRPO with verified reward.}
    \vspace{-2em}
    \label{fig:method}
\end{figure*}

\subsection{Task Formulation}
We consider the visual trace reasoning (VTR) task as shown in~\autoref{fig:method}, 
where a model receives a prompt $c$ and an initial maze image $I_0$, and must generate 
a video $V = \{f_1, f_2, \ldots, f_N\}$ showing an agent (e.g., a colored ball) 
navigating from a start position to a goal position. The task requires the model to 
perceive the maze structure from the initial image, plan a valid path avoiding 
obstacles, and generate temporally consistent motion to the goal.

\subsection{Background: Flow Matching}

Flow matching models~\citep{lipman2023flowmatchinggenerativemodeling,liu2022flowstraightfastlearning} define a continuous transformation between noise and data through a velocity field $v_\theta(x_t, t)$. The forward process interpolates between data $x_0$ and noise $x_1$:
\begin{equation}
    x_t = (1-t)x_0 + tx_1, \quad t \in [0,1]
\end{equation}

The model is trained to predict the velocity field by minimizing:
\begin{equation}
    \mathcal{L}(\theta) = \mathbb{E}_{t, x_0, x_1}\left[\|v - v_\theta(x_t, t)\|^2\right]
\end{equation}

For inference, samples are generated by solving ordinary differential equation (ODE):
\begin{equation}
    dx_t = v_\theta(x_t, t)dt
\end{equation}

\subsection{RL for Flow Matching}

Group Relative Policy Optimization (GRPO)~\citep{guo2025deepseek} provides a lightweight alternative to PPO~\citep{schulman2017proximalpolicyoptimizationalgorithms} by using group-relative advantage estimation, eliminating the need for a separate value network.

\paragraph{ODE to SDE Conversion.}
A key challenge in applying GRPO to flow models is that the deterministic ODE sampling prevents the stochastic exploration required for RL. Flow-GRPO~\citep{liu2025flowgrpotrainingflowmatching} convert the ODE to an equivalent stochastic differential equation (SDE) that preserves the marginal distribution while introducing randomness:
\begin{equation}
    dx_t = \left(v_t(x_t) - \frac{\sigma_t^2}{2}\nabla\log p_t(x_t)\right)dt + \sigma_t dw
\end{equation}

where $\sigma_t$ controls the noise level and $dw$ denotes Wiener process increments. For rectified flow, the discretized update becomes:
\begin{equation}
    x_{t+\Delta t} = x_t + \tilde{v}_\theta(x_t, t)\Delta t + \sigma_t\sqrt{\Delta t}\epsilon
\end{equation}
where $\epsilon \sim \mathcal{N}(0, I)$ and $\tilde{v}_\theta$ includes the score correction term.

\paragraph{Group Relative Advantage.}
Given a prompt-image pair $(c, I_0)$, we sample a group of $G$ videos $\{V^i\}_{i=1}^G$ using SDE sampling. The advantage for each video is computed by normalizing rewards within the group:
\begin{equation}
    \hat{A}^i = \frac{R(V^i, c) - \text{mean}(\{R(V^j, c)\}_{j=1}^G)}{\text{std}(\{R(V^j, c)\}_{j=1}^G)}
\end{equation}

\paragraph{Policy Optimization.}
The GRPO objective maximizes:
\begin{equation}
    \mathcal{J}(\theta) = \mathbb{E}\left[\frac{1}{G}\sum_{i=1}^G\frac{1}{T}\sum_{t=0}^{T-1}\left(\mathcal{L}_{\text{clip}}^{i,t} - \beta_{KL} D_{\text{KL}}\right)\right]
\end{equation}
where $\mathcal{L}_{\text{clip}}^{i,t}$ is the clipped surrogate objective and $\beta_{KL}$ controls KL regularization against the reference policy.

\paragraph{Denoising Reduction.}
Video generation is computationally expensive, requiring many denoising steps per sample. We follow denoising reduction~\citep{liu2025flowgrpotrainingflowmatching}: using fewer steps ($S_{\text{train}}=30$) during training data collection while retaining full steps ($S_{\text{infer}}=50$) during evaluation. This significantly accelerates training without sacrificing final performance.

\subsection{Verifiable Reward Design for Game}

A critical---and underexplored---challenge in applying RL to video 
generation is reward design. Unlike text-based RL where human 
preference provides a natural reward signal, video generation tasks 
present unique difficulties: the output space is high-dimensional and 
continuous, success criteria are often spatial and temporal, and learned 
reward models face severe reward hacking risks specific to the visual 
domain.

We identify three desiderata for effective video RL rewards: 
(i)~\textbf{verifiability}---the reward must reflect genuine task 
success, not superficial visual quality; 
(ii)~\textbf{density}---the reward should provide gradient information 
beyond binary success/failure; and 
(iii)~\textbf{decomposability}---the reward should separate distinct 
aspects of task performance to prevent conflation. 
We now describe how our reward design satisfies each criterion.

\paragraph{Trajectory Extraction.}
Given a generated video, we extract the agent's trajectory using object tracking~\citep{yang2025reasoningvideoevaluationvideo}. We initialize a tracker on the agent's bounding box in the first frame and track its center position across all frames, yielding a trajectory $\tau_{\text{pred}} = \{p_1, p_2, \ldots, p_N\}$. Trajectory samples are presented in Appendix.

\paragraph{Reward Components.}
We define rewards based on trajectory-level metrics~\citep{yang2025reasoningvideoevaluationvideo}. Let $\{\hat{v}_{ij}\}_{j=1}^{n_i}$ denote the predicted trajectory steps and $\{v_{ij}\}_{j=1}^{n_i}$ denote the ground-truth optimal path for the $i$-th sample.

\textbf{Exact Match Reward ($R_{\text{EM}}$):} Measures whether the generated trajectory exactly matches the complete optimal path:
\begin{equation}
    R_{\text{EM}} = \prod_{j=1}^{n_i} \mathbb{I}(\hat{v}_{ij} = v_{ij})
\end{equation}
This is a strict binary reward that requires every step to be correct. One deviation from the optimal solution yields zero reward.

\textbf{Precision Reward ($R_{\text{PR}}$):} Quantifies the proportion of consecutively correct steps along the optimal path, providing a softer metric than exact match:
\begin{equation}
    R_{\text{PR}} = \frac{1}{n_i}\sum_{j=1}^{n_i}\left[\prod_{k=1}^{j}\mathbb{I}(\hat{v}_{ik} = v_{ik})\right]
\end{equation}
This reward reflects the model's ability to make steady, meaningful progress toward the complete correct trajectory.

\textbf{Fidelity Reward ($R_{\text{MF}}$):} Measures the structural consistency of the maze layout across video frames, penalizing unrealistic behaviors such as walls changing or disappearing:
\begin{equation}
    R_{\text{MF}} = \frac{1}{M}\sum_{m=1}^{M}\left(1 - \frac{|\{p : |I_0(p) - I_m(p)| > \tau\}|}{N_m}\right)
\end{equation}
where $M$ is the number of sampled frames, $I_0$ and $I_m$ denote the background regions of the first and $m$-th frames respectively, $\tau$ is a pixel-difference threshold, and $N_m$ is the number of valid overlapping pixels. Higher values indicate better preservation of the static maze structure.

\paragraph{Combined Reward.}
Our final reward combines these components as shown in~\autoref{fig:method}:
\begin{equation}
    R = \alpha R_{\text{EM}} + \beta R_{\text{PR}} + \gamma R_{\text{MF}}
\end{equation}
where $\alpha, \beta, \gamma$ are hyperparameters balancing path correctness, progressive accuracy, and visual consistency. In our experiments, we use $\alpha = 0.3$, $\beta = 0.5$, $\gamma = 0.2$, prioritizing the precision reward which provides denser learning signal while still encouraging exact solutions and physical plausibility.

% \paragraph{Why Verifiable Rewards?}
% Trajectory-based rewards are grounded in 
% objective task success criteria, cannot be ``gamed'' without solving the task, 
% provide interpretable feedback for analysis, and scale reliably to 
% out-of-distribution scenarios.

\subsection{Verifiable Reward Design for Robotic Reasoning}
\label{sec:vggt_reward}
For robotic reasoning task, such as Target-Bench~\citep{wang2025targetbench}, supervision is given as a \emph{reference rollout video} collected by a teleoperated robot, rather than a discrete action sequence. We therefore design a reward that directly compares a generated video plan to its reference, without relying on a learned judge.

\paragraph{Problem setting.}
Each Target-Bench sample includes a semantic goal prompt $c$, a first-frame observation $I_0$, and a reference video $V^{*}=\{f^{*}_1,\ldots,f^{*}_{T^{*}}\}$. Conditioned on $(c,I_0)$, the model generates a video rollout $V=\{f_1,\ldots,f_T\}$ that should depict motion toward the target.

\paragraph{Frame embedding reward.}
We compute per-frame embeddings using a frozen vision encoder $\phi$ (DINOv2~\citep{oquab2023dinov2} or CLIP~\citep{radford2021clip}) and $\ell_2$ normalization:
$e_t=\mathrm{norm}(\phi(f_t))$ and $e^{*}_t=\mathrm{norm}(\phi(f^{*}_t))$.
Since $T$ and $T^{*}$ may differ, we linearly interpolate the shorter embedding sequence to a common length $T_c$.
We then compute (i) mean frame cosine similarity $R_{\text{cos}}$, (ii) a temporal-order consistency score $R_{\text{temp}}$ based on normalized cumulative displacement in embedding space, and (iii) endpoint fidelity $R_{\text{end}}$:
\begin{align}
R_{\text{cos}} &= \frac{1}{T_c}\sum_{t=1}^{T_c}\langle e_t, e_t^{*}\rangle, \\
R_{\text{end}} &= \frac{1}{2}\Big(\langle e_1,e_1^{*}\rangle + \langle e_{T_c}, e_{T_c}^{*}\rangle\Big), \\
R_{\text{temp}} &= 1 - \frac{1}{T_c-1}\sum_{t=1}^{T_c-1}\left|\bar{c}_t - \bar{c}^{*}_t\right|,
\end{align}
where $\bar{c}_t = \frac{\sum_{k=1}^{t}\|e_{k+1}-e_k\|_2}{\sum_{k=1}^{T_c-1}\|e_{k+1}-e_k\|_2 + \epsilon}$ (and similarly $\bar{c}^{*}_t$).
The embedding reward is a weighted sum:
\begin{equation}
R_{\text{emb}} = \alpha_{\text{emb}} R_{\text{cos}} + \beta_{\text{emb}} R_{\text{temp}} + \gamma_{\text{emb}} R_{\text{end}},
\end{equation}
with $(\alpha_{\text{emb}},\beta_{\text{emb}},\gamma_{\text{emb}})=(0.5,0.2,0.3)$ in our implementation.

% \paragraph{Optional VGGT trajectory reward.}
% To incorporate geometry, we could optionally decode camera extrinsics from both $V$ and $V^{*}$ using VGGT~\citep{wang2025vggt}. Following the Target-Bench evaluation protocol~\citep{wang2025targetbench}, we can recover a 2D camera trajectory (with segment-level scale recovery) and compute the weighted overall score $S_{\text{WO}}\in[0,1]$; We can then use $R_{\text{traj}} = S_{\text{WO}}$ as a trajectory reward.

% \paragraph{Combined Target-Bench reward.}
% Our final reward for Target-Bench is:
% \begin{equation}
% R_{\text{TB}} = w_{\text{emb}} R_{\text{emb}} + w_{\text{traj}} R_{\text{traj}}.
% \end{equation}
% We use $w_{\text{emb}}=1.0$ by default, and enable $w_{\text{traj}}>0$ when VGGT inference is available (Sec.~\ref{sec:targetbench_eval}).

% \paragraph{Reward API.}
% For efficiency, we also provide the VGGT reward as a lightweight HTTP service, decoupling RL training from heavy vision backbones and geometry decoding.

\subsection{Training Algorithm}

Algorithm~\ref{alg:video-grpo} summarizes the complete Video-GRPO training procedure.

\begin{algorithm}[h]
\caption{Video-GRPO Training}
\label{alg:video-grpo}
\begin{algorithmic}[1]
\REQUIRE SFT model $\pi_\theta$, reference $\pi_{\text{ref}}$, dataset $\mathcal{D}$, group size $G$, training epoch $T_{\text{train}}$
\FOR{each iteration}
    \STATE Sample batch of $(c, I, \tau^*)$ from $\mathcal{D}$
    \FOR{each $(c, I, \tau^*)$ in batch}
        \STATE Sample $G$ videos via SDE with $S_{\text{train}}$ steps
        \STATE Extract trajectories $\{\hat{\tau}^i\}_{i=1}^G$ via tracking
        \STATE Compute rewards $\{r^i\}_{i=1}^G$ against $\tau^*$
        \STATE Compute advantages $\{\hat{A}^i\}_{i=1}^G$ via Eq. (6)
    \ENDFOR
    \STATE Update $\theta$ by maximizing $\mathcal{J}(\theta)$ via Eq. (7)
\ENDFOR
\RETURN Trained model $\pi_\theta$
\end{algorithmic}
% \vspace{-1em}
\end{algorithm}

% ============================================
% 4. EXPERIMENTAL SETUP
% ============================================
\section{Experimental Setup}

\subsection{Dataset and Tasks}

We conduct experiments on maze-solving tasks from VR-Bench~\citep{yang2025reasoningvideoevaluationvideo}, which provides procedurally generated mazes across five types:
\begin{itemize}
    \item \textbf{Regular Maze:} Grid-based layouts testing basic pathfinding
    \item \textbf{Irregular Maze:} Curved paths preventing coordinate shortcuts
    \item \textbf{3D Maze:} Stereoscopic structures requiring depth perception
    \item \textbf{Trapfield:} Obstacle avoidance with trap regions
    \item \textbf{Sokoban:} Box-pushing puzzles requiring rule comprehension
\end{itemize}

Each maze type includes three difficulty levels (Easy, Medium, Hard) and multiple visual textures. Examples of games are presented in Appendix.

\paragraph{Target-Bench (robotic path planning).}
We additionally include \textbf{Target-Bench}~\citep{wang2025targetbench}, a real-world benchmark for \emph{mapless} path planning toward text-specified \emph{semantic targets}. 
Each sample provides an egocentric RGB observation $I_0$, a goal prompt $c$ (explicit or implicit), and a reference rollout video $V^{*}$ collected with a quadruped robot, together with SLAM-verified ground-truth trajectories. 
We follow the official protocol and evaluate using the VGGT world decoder and trajectory metrics.

\subsection{Base Model}

We use \textit{Wan2.2-TI2V-5B} \citep{wan2025wan} as our base video generation model. This is a flow-matching model for text-and-image-to-video generation. We start from an SFT checkpoint that has been fine-tuned on maze-solving demonstrations, which we refer to as Wan-SFT~\citep{yang2025reasoningvideoevaluationvideo}. We compare our model Wan-R1 against a diverse set of baseline models across two categories: \textbf{\textit{(1)}} Six closed-source video models, including \textit{Veo-3.1-fast}, \textit{Veo-3.1-pro} \citep{GoogleVeo2ModelPage}, \textit{Sora-2} \citep{openai2025sora2}, \textit{Kling-v1} \citep{KlingAI2025}, \textit{Seedance-1.0-pro} \citep{Seedance10Pro}, and \textit{MiniMax-Hailuo-2.3} \citep{MiniMaxHailuo23}; and \textbf{\textit{(2)}} three open-source video models, namely \textit{Wan2.5-i2v-preview} \citep{AlibabaWan25} and \textit{Wan2.2-TI2V-5B} \citep{wan2025wan}, as well the supervised finetuned \textit{Wan-SFT} \citep{yang2025reasoningvideoevaluationvideo}

\subsection{Training Configuration}

The training configurations are presented in Appendix. Besides, we use $\alpha = 0.3$, $\beta = 0.5$, $\gamma = 0.2$ for the combined reward. For Target-Bench, we use the reward described in Sec.~\ref{sec:vggt_reward}. We set $(\alpha_{\text{emb}},\beta_{\text{emb}},\gamma_{\text{emb}})=(0.5,0.2,0.3)$. We use LoRA~\cite{hu2021loralowrankadaptationlarge} to finetune the model. All experiments are conducted on machine with H200 GPUs. 

% \begin{table}[h]
% \centering
% \caption{Training Hyperparameters}
% \label{tab:grpo_hyperparams}
% \begin{tabular}{lc}
% \toprule
% \textbf{Hyperparameter} & \textbf{Value} \\
% \midrule
% Group size & $G = 8$ \\
% LoRA rank & $r = 32$ \\
% Number of frames    & 193 \\
% Training Epoch & $T_{\text{train}} = 1$ \\
% Inference denoising steps & $S_{\text{infer}} = 50$ \\
% KL coefficient & $\beta_{KL} = 0.04$ \\
% Clipping parameter & $\epsilon = 0.2$ \\
% SDE noise scale & $a = 0.5$ \\
% Learning rate & $1 \times 10^{-4}$ \\
% \bottomrule
% \end{tabular}
% \label{tab:config}
% \end{table}

\subsection{Evaluation Metrics}

We evaluate using four metrics from VR-Bench~\citep{yang2025reasoningvideoevaluationvideo}:
\begin{itemize}
    \item \textbf{Exact Match (EM):} Whether generated trajectory exactly matches the optimal path
    \item \textbf{Success Rate (SR):} Whether the agent reaches the goal
    \item \textbf{Precision Rate (PR):} Proportion of consecutively correct steps
    \item \textbf{Step Deviation (SD):} Relative path length redundancy (lower is better)
\end{itemize}

% We also evaluate video quality using \textbf{Maze Fidelity (MF)} which is defined as structural consistency of the maze across frames.
\paragraph{Target-Bench.}

We follow the Target-Bench evaluation protocol~\citep{wang2025targetbench} to assess whether a generated video implies a useful navigation plan. We report five metrics: Average Displacement Error (ADE), Final Displacement Error (FDE), Miss Rate (MR, threshold $\tau{=}2.0$\,m), Soft Endpoint score (SE, tolerance $\sigma{=}0.6$\,m), and Approach Consistency (AC), which measures whether the predicted trajectory stays within a progress-dependent corridor around the ground-truth path.
These are aggregated into a weighted overall score $\text{WO}\in[0,1]$, with AC and SE jointly carrying $65\%$ of the weight.% The pipeline has two stages: (i) a \emph{world decoder} that recovers camera motion from the generated video using VGGT~\citep{wang2025vggt}, with monocular scale ambiguity resolved via ground-truth displacement metadata; and (ii) a \emph{path evaluation} module that projects scaled camera translations onto the ground plane and compares the resulting 2D trajectory against SLAM ground truth.

% ============================================
% 5. RESULTS
% ============================================
\section{Results}

\subsection{Main Results}

\begin{table*}[h]
\vspace{-1em}
\centering
\setlength{\tabcolsep}{0.9mm}{
\resizebox{1.0\linewidth}{!}{
\begin{tabular}{c|l|ccccc|ccccc|ccccc|ccccc}
\toprule
\multicolumn{2}{c|}{\multirow{2}{*}{\textbf{Method}}} &
\multicolumn{5}{c|}{\textbf{EM}~($\uparrow$)} & 
\multicolumn{5}{c|}{\textbf{SR}~($\uparrow$)} & 
\multicolumn{5}{c|}{\textbf{PR}~($\uparrow$)} & 
\multicolumn{5}{c}{\textbf{SD}~($\downarrow$)} \\
\multicolumn{2}{c|}{} & Base & Irreg & Trap & 3D & Soko & Base & Irreg & Trap & 3D & Soko & Base & Irreg & Trap & 3D & Soko & Base & Irreg & Trap & 3D & Soko \\ \midrule
\multirow{11}{*}{\rotatebox{90}{\textbf{General Video Model}}} 
& \multicolumn{20}{c}{\textbf{Closed-Source}} \\[2pt]
& Veo-3.1-fast          & 0.0 & 0.0 & 0.0 & 0.0 & 2.8 & 40.3 & 36.1 & 38.9 & 48.6 & 43.1 & 20.2 & 24.8 & 28.2 & 13.4 & 21.7 & 195.3 & 111.5 & 80.7 & 33.5 & 112.3 \\ 
& Veo-3.1-pro           & 0.0 & 4.2 & 1.4 & 0.0 & 0.0 & 47.2 & 36.1 & 59.7 & 50.0 & 37.5 & 24.6 & 33.9 & 39.1 & 18.0 & 21.4 & 140.7 & 94.5 & 85.4 & 40.1 & 141.8 \\
& Sora-2                & 1.4 & 5.6 & 0.0 & 0.0 & 4.2 & \underline{75.0} & \textbf{72.2} & \underline{83.0} & 37.5 & 43.1 & \underline{45.1} & 45.7 & 46.6 & 19.3 & 27.4 & 302.9 & 187.0 & 145.1 & 92.4 & 138.7 \\ 
% & Sora-2-pro            &  &  &  &  &  &  &  &  &  &  &  &  &  &  &  &  &  &  &  \\ 
& kling-v1              & 0.0 & 0.0 & 0.0 & 0.0 & 0.0 & 2.8 & 0.0 & 1.4 & 27.8 & 12.5 & 6.3 & 8.8 & 10.4 & 11.7 & 9.0 & 25.2 & -- & -- & 69.7 & 356.1 \\
& Seedance-1.0-pro      & 0.0 & 2.8 & 2.8 & 0.0 & 0.0 & 75.0 & 45.8 & 59.7 & \underline{77.8} & 13.9 & 12.8 & 35.8 & 42.7 & 23.6 & 17.1 & 162.3 & 143.4 & 99.1 & 84.4 & 241.9 \\
& MiniMax-Hailuo-2.3    & 0.0 & 1.4 & 2.8 & 0.0 & 0.0 & 68.1 & 40.3 & 70.8 & 55.6 & \underline{45.8} & 23.2 & 24.2 & 30.3 & 20.3 & 15.5 & 464.0 & 170.0 & 90.9 & 50.1 & 165.5 \\ \cline{2-22} \noalign{\vskip 3pt}
% & \multicolumn{1}{c}{\textbf{Open-Source}} & \\
& \multicolumn{20}{c}{\textbf{Open-Source}} \\[2pt]
& Wan2.5-i2v-preview    & 0.0 & 2.8 & 4.2 & 0.0 & 0.0 & 58.3 & 26.4 & 77.8 & 24.5 & 22.4 & 14.3 & 21.8 & 34.4 & 24.5 & 17.1 & 378.4 & 281.8 & 73.2 & 119.9 & 278.0 \\
& Wan2.2-TI2V-5B$^\Diamond$ & 0.0 & 0.0 & 0.0 & 0.0 & 0.0 & 6.9 & 12.5 & 0.0 & 31.9 & 11.1 & 6.6 & 9.1 & 7.1 & 12.8 & 9.2 & 388.7 & 66.1 & -- & 5.4 & 176.6\\ 
& Wan-SFT~\citep{yang2025reasoningvideoevaluationvideo} & 
33.3 & 56.9 & 38.9 & 65.3 & 4.2 & 
76.4 & \underline{69.4} & 100.0 & 100.0 & 69.4 & 
60.6 & 71.6 & 79.1 & 93.5 & 44.3 & 
10.3 & 2.4 & \underline{3.9} & \underline{3.9} & 10.2 \\
\midrule 
\multirow{1}{*}{\rotatebox{90}{\textbf{Train}}} 
& Wan-R1(ours) & 
\cellcolor{yellow!20}{61.1} & \cellcolor{yellow!20}{47.9} & \cellcolor{yellow!20}{90.3} & \cellcolor{yellow!20}{94.4} & \cellcolor{yellow!20}{30.6} & 
\cellcolor{yellow!20}{75.0} & \cellcolor{yellow!20}{69.7} & \cellcolor{yellow!20}{100.0} & \cellcolor{yellow!20}{100.0} & \cellcolor{yellow!20}{68.1} & 
\cellcolor{yellow!20}{78.6} & \cellcolor{yellow!20}{68.4} & \cellcolor{yellow!20}{97.7} & \cellcolor{yellow!20}{98.0} & \cellcolor{yellow!20}{46.9} & 
\cellcolor{yellow!20}{5.2} & \cellcolor{yellow!20}{8.3} & \cellcolor{yellow!20}{2.4} & \cellcolor{yellow!20}{1.9} & \cellcolor{yellow!20}{16.9} \\
& \multicolumn{1}{c|}{$\boldsymbol{\Delta}\!\uparrow$} &
\cellcolor{yellow!20}{+61.1} & \cellcolor{yellow!20}{+47.9} & \cellcolor{yellow!20}{+90.3} & \cellcolor{yellow!20}{+94.4} & \cellcolor{yellow!20}{+30.6} & 
\cellcolor{yellow!20}{+68.1} & \cellcolor{yellow!20}{+57.2} & \cellcolor{yellow!20}{+100.0} & \cellcolor{yellow!20}{+68.1} & \cellcolor{yellow!20}{+57.0} & 
\cellcolor{yellow!20}{+72.0} & \cellcolor{yellow!20}{+59.3} & \cellcolor{yellow!20}{+90.6} & \cellcolor{yellow!20}{+85.2} & \cellcolor{yellow!20}{+37.7} & 
\cellcolor{yellow!20}{-383.5} & \cellcolor{yellow!20}{-57.8} & \cellcolor{yellow!20}{--} & \cellcolor{yellow!20}{-3.5} & \cellcolor{yellow!20}{-159.7} \\
\bottomrule
\end{tabular}}}

\caption{
VR-Bench comprises five tasks: Base (Regular Maze), Irreg (Irregular Maze), Trap (TrapField), 3D (3D Maze), and Soko (Sokoban). Last row shows the performance gain compared with Wan2.2-TI2V base model.
}
\vspace{-3em}
\label{tab:main_results}
\end{table*}

Table~\ref{tab:main_results} presents the main comparison between SFT and RL-tuned models across all maze types. The RL-tuned model (Wan-R1) consistently outperforms the SFT baseline across all maze types and metrics. On Regular Maze, Wan-R1 achieves 61.1\% exact match compared to 33.3\% for Wan-SFT, representing an absolute gain of 27.8 percentage points. For 3D Maze, the improvement is even more pronounced, with exact match increasing from 65.3\% to 94.4\%. Trapfield shows the most dramatic gains, where exact match jumps from 38.9\% to 90.3\%, accompanied by near-perfect precision (97.7\%). While success rates remain comparable across both models for most tasks (as both achieve high completion rates), the key improvements manifest in path optimality: step deviation decreases substantially on Regular Maze (from 10.3 to 5.2) and 3D Maze (from 3.9 to 1.9), indicating that RL training produces more efficient trajectories that closely follow optimal solutions.

\subsection{Generalization Results}

\paragraph{Difficulty Generalization.}

\begin{table}[h]
\centering
\setlength{\tabcolsep}{3pt}
\renewcommand{\arraystretch}{0.85}
\resizebox{\linewidth}{!}{
\footnotesize
\begin{tabular}{l|ccc|ccc|ccc|ccc}
\toprule
\textbf{Task} 
& \multicolumn{3}{c|}{\textbf{EM}} 
& \multicolumn{3}{c|}{\textbf{SR}} 
& \multicolumn{3}{c|}{\textbf{PR}} 
& \multicolumn{3}{c}{\textbf{SD}} \\
& E & M & H & E & M & H & E & M & H & E & M & H \\
\midrule

% ---------- Base (maze) ----------
\multirow{3}{*}{Base} 
& 0.0 & 0.0 & 0.0 
& 8.3 & 8.3 & 4.2
& 13.2 & 3.8 & 2.7
& 154.8 & -- & -- 
\\
& 
\cellcolor{cyan!15}{\scriptsize 0.0\,({\color{gray}+0.0})}
& \cellcolor{cyan!15}{\scriptsize 4.2\,({\color{BetterGreen}+4.2})}
& \cellcolor{cyan!15}{\scriptsize 0.0\,({\color{gray}+0.0})}
& \cellcolor{cyan!15}{\scriptsize 83.3\,({\color{BetterGreen}+75.0})}
& \cellcolor{cyan!15}{\scriptsize 41.7\,({\color{BetterGreen}+33.4})}
& \cellcolor{cyan!15}{\scriptsize 58.3\,({\color{BetterGreen}+54.1})}
& \cellcolor{cyan!15}{\scriptsize 40.1\,({\color{BetterGreen}+26.9})}
& \cellcolor{cyan!15}{\scriptsize 26.1\,({\color{BetterGreen}+22.3})}
& \cellcolor{cyan!15}{\scriptsize 6.0\,({\color{BetterGreen}+3.3})}
& \cellcolor{cyan!15}{\scriptsize 28.2\,({\color{BetterGreen}-126.6})}
& \cellcolor{cyan!15}{\scriptsize 10.1\,({\color{gray}--})}
& \cellcolor{cyan!15}{\scriptsize --\,({\color{gray}--})}
\\
& 
\cellcolor{yellow!20}{\scriptsize 87.5\,({\color{BetterGreen}+87.5})}
& \cellcolor{yellow!20}{\scriptsize 66.7\,({\color{BetterGreen}+66.7})}
& \cellcolor{yellow!20}{\scriptsize 29.2\,({\color{BetterGreen}+29.2})}
& \cellcolor{yellow!20}{\scriptsize 100.0\,({\color{BetterGreen}+91.7})}
& \cellcolor{yellow!20}{\scriptsize 87.5\,({\color{BetterGreen}+79.2})}
& \cellcolor{yellow!20}{\scriptsize 41.7\,({\color{BetterGreen}+37.5})}
& \cellcolor{yellow!20}{\scriptsize 97.0\,({\color{BetterGreen}+83.8})}
& \cellcolor{yellow!20}{\scriptsize 85.4\,({\color{BetterGreen}+81.6})}
& \cellcolor{yellow!20}{\scriptsize 57.2\,({\color{BetterGreen}+54.5})}
& \cellcolor{yellow!20}{\scriptsize 1.4\,({\color{BetterGreen}-153.4})}
& \cellcolor{yellow!20}{\scriptsize 2.5\,({\color{gray}--})}
& \cellcolor{yellow!20}{\scriptsize 14.6\,({\color{gray}--})}
\\
\midrule

% ---------- Irregular ----------
\multirow{3}{*}{Irrg}
& 0.0 & 0.0 & 0.0
& 29.2 & 4.2 & 4.2
& 13.4 & 8.1 & 5.8
& 48.3 & -- & 39.3 
\\
&
\cellcolor{cyan!15}{\scriptsize 83.3\,({\color{BetterGreen}+83.3})}
& \cellcolor{cyan!15}{\scriptsize 66.7\,({\color{BetterGreen}+66.7})}
& \cellcolor{cyan!15}{\scriptsize 54.2\,({\color{BetterGreen}+54.2})}
& \cellcolor{cyan!15}{\scriptsize 95.8\,({\color{BetterGreen}+66.6})}
& \cellcolor{cyan!15}{\scriptsize 87.5\,({\color{BetterGreen}+83.3})}
& \cellcolor{cyan!15}{\scriptsize 62.5\,({\color{BetterGreen}+58.3})}
& \cellcolor{cyan!15}{\scriptsize 88.0\,({\color{BetterGreen}+74.6})}
& \cellcolor{cyan!15}{\scriptsize 74.8\,({\color{BetterGreen}+66.7})}
& \cellcolor{cyan!15}{\scriptsize 68.4\,({\color{BetterGreen}+62.6})}
& \cellcolor{cyan!15}{\scriptsize 3.5\,({\color{BetterGreen}-44.8})}
& \cellcolor{cyan!15}{\scriptsize 7.8\,({\color{gray}--})}
& \cellcolor{cyan!15}{\scriptsize 3.1\,({\color{BetterGreen}-36.2})}
\\
&
\cellcolor{yellow!20}{\scriptsize 79.2\,({\color{BetterGreen}+79.2})}
& \cellcolor{yellow!20}{\scriptsize 45.8\,({\color{BetterGreen}+45.8})}
& \cellcolor{yellow!20}{\scriptsize 20.8\,({\color{BetterGreen}+20.8})}
& \cellcolor{yellow!20}{\scriptsize 100.0\,({\color{BetterGreen}+70.8})}
& \cellcolor{yellow!20}{\scriptsize 66.7\,({\color{BetterGreen}+62.5})}
& \cellcolor{yellow!20}{\scriptsize 37.5\,({\color{BetterGreen}+33.3})}
& \cellcolor{yellow!20}{\scriptsize 82.3\,({\color{BetterGreen}+68.9})}
& \cellcolor{yellow!20}{\scriptsize 70.7\,({\color{BetterGreen}+62.6})}
& \cellcolor{yellow!20}{\scriptsize 51.6\,({\color{BetterGreen}+45.8})}
& \cellcolor{yellow!20}{\scriptsize 10.1\,({\color{BetterGreen}-38.2})}
& \cellcolor{yellow!20}{\scriptsize 9.1\,({\color{gray}--})}
& \cellcolor{yellow!20}{\scriptsize 4.3\,({\color{BetterGreen}-35.0})}
\\
\midrule

% ---------- Trapfield ----------
\multirow{3}{*}{Trap}
& 0.0 & 0.0 & 0.0
& 0.0 & 0.0 & 0.0
& 6.0 & 6.4 & 8.9
& -- & -- & --
\\
&
\cellcolor{cyan!15}{\scriptsize 62.5\,({\color{BetterGreen}+62.5})}
& \cellcolor{cyan!15}{\scriptsize 0.0\,({\color{gray}+0.0})}
& \cellcolor{cyan!15}{\scriptsize 12.5\,({\color{BetterGreen}+12.5})}
& \cellcolor{cyan!15}{\scriptsize 100.0\,({\color{BetterGreen}+100.0})}
& \cellcolor{cyan!15}{\scriptsize 95.8\,({\color{BetterGreen}+95.8})}
& \cellcolor{cyan!15}{\scriptsize 62.5\,({\color{BetterGreen}+62.5})}
& \cellcolor{cyan!15}{\scriptsize 86.7\,({\color{BetterGreen}+80.7})}
& \cellcolor{cyan!15}{\scriptsize 43.7\,({\color{BetterGreen}+37.3})}
& \cellcolor{cyan!15}{\scriptsize 56.7\,({\color{BetterGreen}+47.8})}
& \cellcolor{cyan!15}{\scriptsize 2.1\,({\color{gray}--})}
& \cellcolor{cyan!15}{\scriptsize 5.9\,({\color{gray}--})}
& \cellcolor{cyan!15}{\scriptsize 3.5\,({\color{gray}--})}
\\
&
\cellcolor{yellow!20}{\scriptsize 79.2\,({\color{BetterGreen}+79.2})}
& \cellcolor{yellow!20}{\scriptsize 100.0\,({\color{BetterGreen}+100.0})}
& \cellcolor{yellow!20}{\scriptsize 75.9\,({\color{BetterGreen}+75.9})}
& \cellcolor{yellow!20}{\scriptsize 100.0\,({\color{BetterGreen}+100.0})}
& \cellcolor{yellow!20}{\scriptsize 100.0\,({\color{BetterGreen}+100.0})}
& \cellcolor{yellow!20}{\scriptsize 100.0\,({\color{BetterGreen}+100.0})}
& \cellcolor{yellow!20}{\scriptsize 94.7\,({\color{BetterGreen}+88.7})}
& \cellcolor{yellow!20}{\scriptsize 100.0\,({\color{BetterGreen}+93.6})}
& \cellcolor{yellow!20}{\scriptsize 86.4\,({\color{BetterGreen}+77.5})}
& \cellcolor{yellow!20}{\scriptsize 4.9\,({\color{gray}--})}
& \cellcolor{yellow!20}{\scriptsize 1.2\,({\color{gray}--})}
& \cellcolor{yellow!20}{\scriptsize 1.0\,({\color{gray}--})}
\\
\midrule

% ---------- 3D Maze ----------
\multirow{3}{*}{3D}
& 0.0 & 0.0 & 0.0
& 54.2 & 16.7 & 25.0
& 7.6 & 9.8 & 15.2
& 53.0 & 87.9 & 33.6
\\
&
\cellcolor{cyan!15}{\scriptsize 41.7\,({\color{BetterGreen}+41.7})}
& \cellcolor{cyan!15}{\scriptsize 0.0\,({\color{gray}+0.0})}
& \cellcolor{cyan!15}{\scriptsize 4.2\,({\color{BetterGreen}+4.2})}
& \cellcolor{cyan!15}{\scriptsize 100.0\,({\color{BetterGreen}+45.8})}
& \cellcolor{cyan!15}{\scriptsize 79.2\,({\color{BetterGreen}+62.5})}
& \cellcolor{cyan!15}{\scriptsize 83.3\,({\color{BetterGreen}+58.3})}
& \cellcolor{cyan!15}{\scriptsize 78.7\,({\color{BetterGreen}+71.1})}
& \cellcolor{cyan!15}{\scriptsize 47.4\,({\color{BetterGreen}+37.6})}
& \cellcolor{cyan!15}{\scriptsize 58.7\,({\color{BetterGreen}+43.5})}
& \cellcolor{cyan!15}{\scriptsize 4.6\,({\color{BetterGreen}-48.4})}
& \cellcolor{cyan!15}{\scriptsize 10.2\,({\color{BetterGreen}-77.7})}
& \cellcolor{cyan!15}{\scriptsize 13.8\,({\color{BetterGreen}-19.8})}
\\
&
\cellcolor{yellow!20}{\scriptsize 87.5\,({\color{BetterGreen}+87.5})}
& \cellcolor{yellow!20}{\scriptsize 95.8\,({\color{BetterGreen}+95.8})}
& \cellcolor{yellow!20}{\scriptsize 91.7\,({\color{BetterGreen}+91.7})}
& \cellcolor{yellow!20}{\scriptsize 100.0\,({\color{BetterGreen}+45.8})}
& \cellcolor{yellow!20}{\scriptsize 100.0\,({\color{BetterGreen}+83.3})}
& \cellcolor{yellow!20}{\scriptsize 100.0\,({\color{BetterGreen}+75.0})}
& \cellcolor{yellow!20}{\scriptsize 98.4\,({\color{BetterGreen}+90.8})}
& \cellcolor{yellow!20}{\scriptsize 99.1\,({\color{BetterGreen}+89.3})}
& \cellcolor{yellow!20}{\scriptsize 96.2\,({\color{BetterGreen}+81.0})}
& \cellcolor{yellow!20}{\scriptsize 1.7\,({\color{BetterGreen}-51.3})}
& \cellcolor{yellow!20}{\scriptsize 1.6\,({\color{BetterGreen}-86.3})}
& \cellcolor{yellow!20}{\scriptsize 2.7\,({\color{BetterGreen}-30.9})}
\\
\midrule

% ---------- Sokoban ----------
\multirow{3}{*}{Soko}
& 0.0 & 0.0 & 0.0
& 20.8 & 8.3 & 4.2
& 14.6 & 8.3 & 5.6
& 354.2 & 61.4 & --
\\
&
\cellcolor{cyan!15}{\scriptsize 4.2\,({\color{BetterGreen}+4.2})}
& \cellcolor{cyan!15}{\scriptsize 0.0\,({\color{gray}+0.0})}
& \cellcolor{cyan!15}{\scriptsize 0.0\,({\color{gray}+0.0})}
& \cellcolor{cyan!15}{\scriptsize 83.3\,({\color{BetterGreen}+62.5})}
& \cellcolor{cyan!15}{\scriptsize 45.8\,({\color{BetterGreen}+37.5})}
& \cellcolor{cyan!15}{\scriptsize 33.3\,({\color{BetterGreen}+29.1})}
& \cellcolor{cyan!15}{\scriptsize 62.2\,({\color{BetterGreen}+47.6})}
& \cellcolor{cyan!15}{\scriptsize 12.5\,({\color{BetterGreen}+4.2})}
& \cellcolor{cyan!15}{\scriptsize 10.4\,({\color{BetterGreen}+4.8})}
& \cellcolor{cyan!15}{\scriptsize 18.8\,({\color{BetterGreen}-335.4})}
& \cellcolor{cyan!15}{\scriptsize 84.5\,({\color{red}+23.1})}
& \cellcolor{cyan!15}{\scriptsize 16.1\,({\color{gray}--})}
\\
&
\cellcolor{yellow!20}{\scriptsize 75.0\,({\color{BetterGreen}+75.0})}
& \cellcolor{yellow!20}{\scriptsize 12.5\,({\color{BetterGreen}+12.5})}
& \cellcolor{yellow!20}{\scriptsize 4.2\,({\color{BetterGreen}+4.2})}
& \cellcolor{yellow!20}{\scriptsize 91.7\,({\color{BetterGreen}+70.9})}
& \cellcolor{yellow!20}{\scriptsize 91.7\,({\color{BetterGreen}+83.4})}
& \cellcolor{yellow!20}{\scriptsize 45.8\,({\color{BetterGreen}+41.6})}
& \cellcolor{yellow!20}{\scriptsize 87.8\,({\color{BetterGreen}+73.2})}
& \cellcolor{yellow!20}{\scriptsize 30.3\,({\color{BetterGreen}+22.0})}
& \cellcolor{yellow!20}{\scriptsize 19.4\,({\color{BetterGreen}+13.8})}
& \cellcolor{yellow!20}{\scriptsize 5.8\,({\color{BetterGreen}-348.4})}
& \cellcolor{yellow!20}{\scriptsize 8.8\,({\color{BetterGreen}-52.6})}
& \cellcolor{yellow!20}{\scriptsize 38.6\,({\color{gray}--})}
\\

\bottomrule
\end{tabular}}

\caption{
Difficulty generalization performance of Wan-R1 on VR-Bench. Each task block presents a comparison between the baseline model (Wan2.2-TI2V-5B) and \colorbox{cyan!15}{Wan-SFT} (trained exclusively on Easy-level data) and \colorbox{yellow!20}{Wan-R1} (trained on easy tasks with flow-GRPO) evaluated across three difficulty levels (E/M/H = Easy/Medium/Hard).}
\vspace{-2em}

\label{tab:difficulty_gen}
\end{table}

Table~\ref{tab:difficulty_gen} shows generalization to harder difficulty levels when trained only on Easy mazes. The GRPO model shows significantly better generalization to Medium and Hard difficulties. On Regular Maze, while the SFT model degrades catastrophically from Easy to Hard (40.1\% to 6.0\% PR), Wan-R1 maintains substantially higher performance (97.0\% to 57.2\% PR). For Irregular Maze, the RL-tuned model achieves 68.4\% precision on Hard difficulty compared to just 5.8\% for the baseline. The pattern holds across most maze types: on 3D Maze Hard, Wan-R1 reaches 96.2\% precision versus 15.2\% baseline, representing a gain of over 80 percentage points. These results demonstrate that RL encourages learning of more generalizable reasoning strategies rather than memorizing specific maze.

\paragraph{Texture Generalization.}

Our appendix presents results on unseen visual textures. The RL-tuned model shows substantially better texture generalization, particularly on visual styles that differ most from training. On Regular Maze, Wan-R1 achieves 31.9\% exact match on both Skin2 and Skin3 compared to 1.4\% and 23.6\% for Wan-SFT respectively. For 3D Maze, the model maintains 79.2\% exact match on unseen textures compared to 43.1\% and 50.0\% for the SFT baseline. This texture invariance suggests that RL encourages the model to focus on structural maze features rather than superficial visual patterns, learning more abstract reasoning strategies that transfer across visual domains.

\paragraph{Maze Type Generalization.}

Our appendix shows cross-task generalization when trained on a single maze type. While fine-tuning on individual tasks improves in-domain performance, our method further enhances both in-domain accuracy and cross-task transfer. For Regular Maze, our approach nearly doubles the in-domain EM (61.1 vs 33.3) while maintaining competitive cross-domain performance. The most substantial gains appear in 3D Maze, where our method achieves 94.4 EM compared to 65.3 from fine-tuning alone, alongside near-perfect PR (98.0). For Sokoban, the inherently more complex task requiring box-pushing logic, our method improves in-domain EM from 4.2 to 30.6 and dramatically boosts cross-domain SR on Base mazes (48.6 vs 0.5). Notably, Trapfield with our method achieves perfect SR (100.0) across Base, Irregular, and 3D domains, demonstrating improved generalization to visually distinct maze types. These results suggest that our approach not only strengthens task-specific learning but also facilitates more robust feature representations that transfer across maze variants.

\subsection{Target-Bench: Mapless Path Planning}

\begin{figure}[h]
\vspace{-2em}
\centering
\includegraphics[width=0.75\columnwidth]{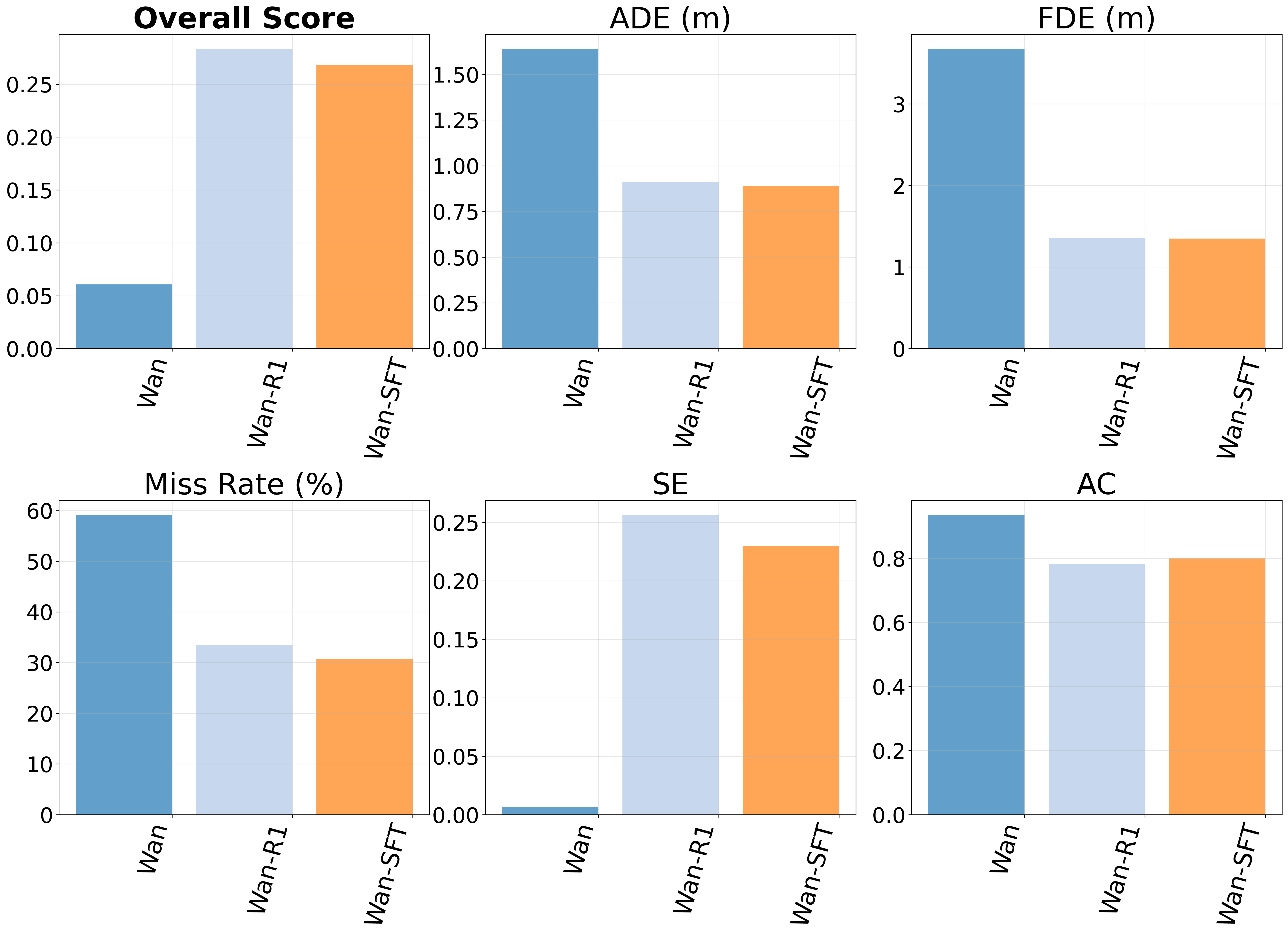} 
\vspace{-1em}
\caption{Target-Bench results. Wan-R1 achieves the 
highest overall score, with lower displacement errors (ADE, FDE) and miss 
rate, and higher soft endpoint (SE) and approach consistency (AC) compared 
to both the base model and Wan-SFT.}
\label{fig:targetbench}
\vspace{-1.5em}
\end{figure}

We additionally evaluate our models on Target-Bench~\citep{wang2025targetbench}, which measures whether a generated video rollout supports mapless navigation toward a semantic target. Figure~\ref{fig:targetbench} provides the results. Wan-R1 achieves the best overall score among all three models, reducing ADE 
and FDE by roughly half relative to the base model while substantially 
lowering miss rate. Both SE and AC also improve, indicating that RL training 
encourages the generated rollouts to better encode goal-directed motion. 
These gains suggest that the embedding-level verifiable reward transfers 
effectively to real-world robotic navigation beyond structured maze tasks.

% ============================================
% 6. ANALYSIS
% ============================================
\section{Analysis}

\subsection{Effect of KL Regularization and Denoising Reduction}

\begin{table}[h]
% \vspace{-2em}
\centering
\small
\begin{subtable}[t]{0.48\textwidth}
\centering
\begin{tabular}{l|cccc}
\toprule
$\beta$ & EM & SR & PR & SD \\
\midrule
0    & 62.5 & 75.0 & 79.0 & 6.7 \\
0.04 & 61.1 & 75.0 & 78.6 & 5.2 \\
0.1  & 66.7 & 75.0 & 81.5 & 3.7 \\
\bottomrule
\end{tabular}
\caption{Effect of KL coefficient $\beta_{KL}$.}
\label{tab:kl_ablation}
\end{subtable}%
\hfill
\begin{subtable}[t]{0.48\textwidth}
\centering
\begin{tabular}{l|cccc}
\toprule
$S_{\text{train}}$ & EM & SR & PR & SD \\
\midrule
5  & 59.7 & 70.8 & 77.5 & 4.0 \\
30 & 61.1 & 75.0 & 78.6 & 5.2 \\
50 & 62.5 & 73.6 & 79.7 & 2.9 \\
\bottomrule
\end{tabular}
\caption{Effect of training denoising steps ($S_{\text{train}}$).}
\label{tab:denoise_ablation}
\end{subtable}
\caption{Ablation studies on KL regularization and denoising reduction.}
\vspace{-3em}
\label{tab:ablation_combined}
\end{table}

\paragraph{KL Regularization.} Table~\ref{tab:kl_ablation} presents the effect of KL 
regularization strength on model performance. Without KL regularization ($\beta_{KL}=0$), the 
model achieves moderate exact match (62.5\%) but exhibits the highest step deviation (6.7), 
indicating a tendency to produce unnecessarily long or redundant paths. Increasing regularization 
to $\beta_{KL}=0.1$ yields the best overall performance, achieving the highest exact match 
(66.7\%), precision rate (81.5\%), and the lowest step deviation (3.7), while maintaining an 
identical success rate (75.0\%) across all settings. This suggests that stronger KL regularization 
encourages the policy to stay closer to the reference distribution, producing more efficient and 
structurally consistent trajectories rather than exploring degenerate solution paths.  

\paragraph{Denoising Reduction.} Reducing training denoising steps dramatically accelerates 
training with minimal performance loss. As shown in Table~\ref{tab:denoise_ablation}, using 
$S_{\text{train}}=5$ degrades performance noticeably (59.7\% EM, 70.8\% SR), confirming that 
an excessively aggressive reduction impairs sample quality during rollout collection. However, 
$S_{\text{train}}=30$ recovers nearly all of the performance of the full 50-step setting 
(61.1\% vs.\ 62.5\% EM, 78.6\% vs.\ 79.7\% PR). Notably, $S_{\text{train}}=30$ achieves the highest 
success rate (75.0\% vs.\ 73.6\% for full steps), suggesting that moderate denoising reduction 
may act as a mild regularizer by introducing controlled stochasticity into the generated rollouts.

\subsection{Cold Start}

Applying Flow-GRPO directly to the base model (Wan2.2-TI2V-5B) without prior supervised fine-tuning yields poor results, as the base model achieves near-zero exact match across all maze types. This highlights a standard characteristic of reinforcement learning: the policy requires a foundational task understanding before RL can effectively guide exploration~\citep{yue2025doesreinforcementlearningreally,feng2025videor1reinforcingvideoreasoning}. Because our verifiable rewards enforce strict logical correctness, they are naturally sparser than subjective visual quality metrics. Therefore, SFT acts as a necessary bootstrap, placing the model within a region of the reward landscape where policy gradient updates become meaningful. Future work could bridge this gap through curriculum learning on simplified environments or offline preference optimization.

\subsection{Test-Time Scaling}

Appendix examines whether RL genuinely improves model capability or merely benefits from best-of-K sampling. We evaluate Wan-R1 with varying numbers of samples $K \in \{1, 4, 8, 12, 16\}$ across difficulty levels. The results reveal that performance continues to improve with larger K, demonstrating that the model maintains meaningful diversity in its generations. Importantly, even at K=1 (no sampling advantage), Wan-R1 substantially outperforms the SFT baseline, confirming that RL training improves the underlying policy rather than simply increasing variance for lucky samples. The scaling curves show diminishing returns beyond K=8, suggesting this as a practical inference budget.

\subsection{Reward Design Analysis}
\label{sec:reward_analysis}

We conduct a systematic study of reward design choices. This analysis serves two purposes: validating our 
multi-component verifiable reward and providing general insights on reward design in video reasoning. The results are presented in Appendix.

\paragraph{Reward models lead to reward hacking.}
We use Qwen2.5-VL as a reward model, prompting it 
to assess whether the generated video correctly solves the maze. 
Appendix reveal fundamental 
failure mode: the model learns to generate videos where the agent moves confidently along plausible-looking but incorrect paths. 
Figure in Appendix illustrates representative examples. The VLM assigns high scores to these incorrect solutions because it 
responds to visual cues of purposeful motion---smooth trajectories, 
clear agent visibility, and goal-directed movement---rather than 
verifying actual path correctness. This confirms that the 
high-dimensional, continuous nature of video outputs makes learned 
reward models particularly vulnerable to exploitation, more so than in text-based RL where outputs are discrete and more easily verified.

% \begin{figure}[t]
% \centering
% % TODO: Add actual figure
% % \includegraphics[width=\linewidth]{figures/reward_hacking.pdf}
% \caption{Reward hacking with VLM-as-Judge. The VLM-rewarded model 
% generates visually confident but incorrect trajectories (middle), 
% receiving high VLM scores despite failing the task. Our verifiable 
% reward (right) produces correct solutions. Numbers below each video 
% show VLM score / trajectory EM.}
% \label{fig:reward_hacking}
% \end{figure}

%--- NEW: Sparse vs. dense reward analysis ---

\paragraph{Sparse rewards provide insufficient learning signal.}
Using only binary success (SR-Only) or exact match ($R_\text{EM}$) as 
reward yields limited improvement over SFT. With SR-Only, the model learns to 
reach the goal but via highly suboptimal routes (high SD), because all 
successful trajectories receive identical reward regardless of 
efficiency. With $R_\text{EM}$ alone, the reward is even sparser---few 
generated trajectories exactly match the optimal path, so most samples 
receive zero reward, providing negligible gradient signal for early 
training.

\paragraph{Dense verifiable rewards enable stable learning.}
The precision reward $R_\text{PR}$ provides the critical dense signal: 
it rewards partial progress along the correct path, creating a smooth 
reward landscape that guides optimization. Training with $R_\text{PR}$ 
alone already achieves better EM, substantially outperforming sparse 
alternatives. Adding $R_\text{EM}$ provides a bonus for complete 
solutions, encouraging the model to ``close out'' trajectories rather 
than settling for partial correctness. The fidelity term $R_\text{MF}$ 
prevents a subtle failure mode where the model achieves high trajectory 
scores by altering the maze structure in later frames (e.g., removing 
walls), which constitutes a form of visual ``cheating.''

% %--- NEW: Training dynamics analysis ---

% \paragraph{Reward component dynamics during training.}

% % TODO: This paragraph requires a training curve figure
% % Plot: x-axis = training step, y-axis = each reward component value
% % Show how R_EM, R_PR, R_MF evolve during training for the Full model
% % This gives reviewers insight into learning dynamics

% Figure~\ref{fig:reward_curves} shows the evolution of each reward 
% component during training. $R_\text{PR}$ improves rapidly in early 
% training as the model learns basic pathfinding, providing consistent 
% gradient signal. $R_\text{EM}$ increases more slowly, reflecting 
% its all-or-nothing nature, but begins to rise once $R_\text{PR}$ 
% plateaus---suggesting that the dense precision signal bootstraps 
% learning that eventually enables exact solutions. $R_\text{MF}$ 
% remains relatively stable, indicating that structural fidelity is 
% maintained throughout training; however, removing it causes a gradual 
% degradation in maze consistency (see $R_\text{PR} + R_\text{MF}$ vs. 
% $R_\text{PR}$ only in Table~\ref{tab:reward_ablation}).

% \begin{figure}[t]
% \centering
% % TODO: Add actual figure
% % \includegraphics[width=0.8\linewidth]{figures/reward_curves.pdf}
% \caption{Training dynamics of reward components. The precision reward 
% $R_\text{PR}$ provides early learning signal, bootstrapping exact 
% match $R_\text{EM}$ improvement in later stages. Fidelity 
% $R_\text{MF}$ remains stable, confirming structural consistency 
% is preserved.}
% \label{fig:reward_curves}
% \end{figure}

\subsection{Why Does RL Improve Generalization?}

We hypothesize several factors contribute to RL's generalization benefits:

\paragraph{Exploration discovers diverse strategies.} Unlike SFT which imitates a single demonstration per maze, RL explores multiple trajectories through stochastic SDE sampling. During training, the model generates G=8 different video solutions for each maze, receiving differentiated feedback based on trajectory quality. This exposure to a broader distribution of valid (and invalid) solutions helps the model learn which features of a solution are essential versus incidental.

\paragraph{Reward provides task-level signal.} SFT optimizes frame-level reconstruction loss, which may cause overfitting to superficial visual patterns such as specific texture details or agent appearances. RL optimizes task completion through trajectory-level rewards, encouraging the model to learn more abstract reasoning strategies that capture the essence of pathfinding rather than pixel-level imitation.

\paragraph{Group normalization reduces spurious correlations.} By comparing within groups of videos generated for the same maze, GRPO's advantage estimation naturally controls for maze-specific factors. A video is considered ``good'' only if it outperforms other attempts on the same maze, not if it happens to be generated for an easier maze. This relative evaluation helps the model learn transferable skills rather than maze-specific shortcuts.

% ============================================
% 7. CONCLUSION
% ============================================
\section{Conclusion}
We explored the application of reinforcement learning to enhance reasoning capabilities in video generation models. By adapting GRPO to flow-based video models and designing verifiable rewards grounded in trajectory matching, we achieved significant improvements in generalization across difficulty levels, visual textures, and maze types. Beyond structured game environments, we further demonstrated that embedding-level verifiable rewards enable effective RL training for robotic video reasoning, where our model improves mapless path planning toward semantic targets on Target-Bench. Our results demonstrate that online RL with carefully designed rewards can strengthen reasoning capabilities across both synthetic and real-world navigation tasks without sacrificing video quality or diversity.

Our work highlights the critical importance of reward design in RL for generative models. While multimodal reward models offer convenience, they are prone to reward hacking in video generation tasks. In contrast, verifiable rewards—whether based on discrete trajectory matching for games or continuous embedding similarity for robotic rollouts—provide more reliable training signals and lead to genuine capability improvements.

\section{Limitations}
Our study focuses on reasoning tasks with strictly definable success criteria, such as ground-truth optimal paths and reference robotic rollouts. While extending this approach to open-ended generation is challenging, this constraint is by design. Evaluating complex spatial and temporal reasoning requires absolute correctness; current vision-language models frequently fail to act as reliable judges in these scenarios, often rewarding visually plausible but logically flawed trajectories. Therefore, grounding video RL in objective, verifiable metrics is a necessary prerequisite for developing models that genuinely reason rather than hallucinate. Additionally, our verifiable rewards are implemented as a weighted combination of multiple components. While our selected fixed weights $(\alpha, \beta, \gamma)$ proved effective and robust in our experiments, the optimal trade-off between path efficiency and visual fidelity may naturally shift depending on the environment's complexity. Future work should explore adaptive or curriculum-based reward weighting to automatically balance these signals during training
 
\newpage
\appendix

\section{Training Configuration}

The training configurations are presented in~\autoref{tab:config}. Additionally, we use $\alpha = 0.3$, $\beta = 0.5$, $\gamma = 0.2$ for the combined reward. For Target-Bench, we use our customized reward which compares generated rollouts against reference videos via (i) frame-level embedding similarity (DINOv2/CLIP) and (ii) optional VGGT-based trajectory scoring. 
Unless stated otherwise, we set $(\alpha_{\text{emb}},\beta_{\text{emb}},\gamma_{\text{emb}})=(0.5,0.2,0.3)$ and use $(w_{\text{emb}},w_{\text{traj}})=(1.0,0.0)$, enabling the trajectory term only when VGGT inference is available. We use LoRA~\cite{hu2021loralowrankadaptationlarge} to finetune the model. All experiments are conducted on a machine with H200 GPUs. 

\begin{table}[h]
\centering
\caption{Training Hyperparameters}
\label{tab:grpo_hyperparams}
\begin{tabular}{lc}
\toprule
\textbf{Hyperparameter} & \textbf{Value} \\
\midrule
Group size & $G = 8$ \\
LoRA rank & $r = 32$ \\
Number of frames    & 193 \\
Training Epoch & $T_{\text{train}} = 1$ \\
Inference denoising steps & $S_{\text{infer}} = 50$ \\
KL coefficient & $\beta_{KL} = 0.04$ \\
Clipping parameter & $\epsilon = 0.2$ \\
SDE noise scale & $a = 0.5$ \\
Learning rate & $1 \times 10^{-4}$ \\
\bottomrule
\end{tabular}
\label{tab:config}
\end{table}

\section{More results}
\subsection{Maze Type Generalization.}
 
\begin{table*}[h]
\centering
\resizebox{\textwidth}{!}{
\begin{tabular}{ll|cccc|cccc|cccc|cccc}
\toprule
\textbf{Task} & \textbf{Model} 
& \multicolumn{4}{c|}{\textbf{EM}} 
& \multicolumn{4}{c|}{\textbf{SR}} 
& \multicolumn{4}{c|}{\textbf{PR}} 
& \multicolumn{4}{c}{\textbf{SD}} \\
& & Base & Irreg & 3D & Soko 
  & Base & Irreg & 3D & Soko 
  & Base & Irreg & 3D & Soko 
  & Base & Irreg & 3D & Soko \\
\midrule

\multicolumn{2}{l|}{\textbf{Wan2.2-TI2V-5B Baseline}}  
& 0.0 & 0.0 & 0.0 & 0.0   
& 6.9 & 12.5 & 31.9 & 11.1   
& 6.6 & 9.1 & 12.8 & 9.2   
& 388.7 & 66.1 & 5.4 & 176.6 \\
\midrule

% ---------- Regular Maze ----------
\multirow{2}{*}{Regular Maze} 
& Wan-SFT 
& \makecell[c]{33.3 \\ \textcolor{BetterGreen}{\scriptsize (+33.3)}} 
& \makecell[c]{5.6 \\ \textcolor{BetterGreen}{\scriptsize (+5.6)}} 
& \makecell[c]{0.0 \\ \textcolor{gray}{\scriptsize (+0.0)}} 
& \makecell[c]{0.0 \\ \textcolor{gray}{\scriptsize (+0.0)}} 
& \makecell[c]{76.4 \\ \textcolor{BetterGreen}{\scriptsize (+69.5)}} 
& \makecell[c]{8.3 \\ \textcolor{red}{\scriptsize (-4.2)}} 
& \makecell[c]{69.4 \\ \textcolor{BetterGreen}{\scriptsize (+37.5)}} 
& \makecell[c]{30.6 \\ \textcolor{BetterGreen}{\scriptsize (+19.5)}} 
& \makecell[c]{60.6 \\ \textcolor{BetterGreen}{\scriptsize (+54.0)}} 
& \makecell[c]{22.7 \\ \textcolor{BetterGreen}{\scriptsize (+13.6)}} 
& \makecell[c]{13.7 \\ \textcolor{BetterGreen}{\scriptsize (+0.9)}} 
& \makecell[c]{19.0 \\ \textcolor{BetterGreen}{\scriptsize (+9.8)}} 
& \makecell[c]{10.3 \\ \textcolor{BetterGreen}{\scriptsize (-378.4)}} 
& \makecell[c]{51.7 \\ \textcolor{BetterGreen}{\scriptsize (-14.4)}} 
& \makecell[c]{12.7 \\ \textcolor{red}{\scriptsize (+7.3)}} 
& \makecell[c]{49.3 \\ \textcolor{BetterGreen}{\scriptsize (-127.3)}} 
\\
& Wan-R1 
& \makecell[c]{61.1 \\ \textcolor{BetterGreen}{\scriptsize (+61.1)}} 
& \makecell[c]{4.2 \\ \textcolor{BetterGreen}{\scriptsize (+4.2)}} 
& \makecell[c]{0.0 \\ \textcolor{gray}{\scriptsize (+0.0)}} 
& \makecell[c]{0.0 \\ \textcolor{gray}{\scriptsize (+0.0)}} 
& \makecell[c]{75.0 \\ \textcolor{BetterGreen}{\scriptsize (+68.1)}} 
& \makecell[c]{13.7 \\ \textcolor{BetterGreen}{\scriptsize (+1.2)}} 
& \makecell[c]{63.9 \\ \textcolor{BetterGreen}{\scriptsize (+32.0)}} 
& \makecell[c]{15.3 \\ \textcolor{BetterGreen}{\scriptsize (+4.2)}} 
& \makecell[c]{78.6 \\ \textcolor{BetterGreen}{\scriptsize (+72.0)}} 
& \makecell[c]{23.9 \\ \textcolor{BetterGreen}{\scriptsize (+14.8)}} 
& \makecell[c]{13.1 \\ \textcolor{BetterGreen}{\scriptsize (+0.3)}} 
& \makecell[c]{20.6 \\ \textcolor{BetterGreen}{\scriptsize (+11.4)}} 
& \makecell[c]{5.2 \\ \textcolor{BetterGreen}{\scriptsize (-383.5)}} 
& \makecell[c]{27.1 \\ \textcolor{BetterGreen}{\scriptsize (-39.0)}} 
& \makecell[c]{10.9 \\ \textcolor{red}{\scriptsize (+5.5)}} 
& \makecell[c]{9.0 \\ \textcolor{BetterGreen}{\scriptsize (-167.6)}} 
\\
\midrule

% ---------- Irregular Maze ----------
\multirow{2}{*}{Irregular Maze} 
& Wan-SFT 
& \makecell[c]{0.0 \\ \textcolor{gray}{\scriptsize (+0.0)}} 
& \makecell[c]{56.9 \\ \textcolor{BetterGreen}{\scriptsize (+56.9)}} 
& \makecell[c]{0.0 \\ \textcolor{gray}{\scriptsize (+0.0)}} 
& \makecell[c]{0.0 \\ \textcolor{gray}{\scriptsize (+0.0)}} 
& \makecell[c]{11.1 \\ \textcolor{BetterGreen}{\scriptsize (+4.2)}} 
& \makecell[c]{69.4 \\ \textcolor{BetterGreen}{\scriptsize (+56.9)}} 
& \makecell[c]{79.2 \\ \textcolor{BetterGreen}{\scriptsize (+47.3)}} 
& \makecell[c]{12.5 \\ \textcolor{BetterGreen}{\scriptsize (+1.4)}} 
& \makecell[c]{16.6 \\ \textcolor{BetterGreen}{\scriptsize (+10.0)}} 
& \makecell[c]{71.6 \\ \textcolor{BetterGreen}{\scriptsize (+62.5)}} 
& \makecell[c]{16.8 \\ \textcolor{BetterGreen}{\scriptsize (+4.0)}} 
& \makecell[c]{15.5 \\ \textcolor{BetterGreen}{\scriptsize (+6.3)}} 
& \makecell[c]{35.8 \\ \textcolor{BetterGreen}{\scriptsize (-352.9)}} 
& \makecell[c]{2.4 \\ \textcolor{BetterGreen}{\scriptsize (-63.7)}} 
& \makecell[c]{9.0 \\ \textcolor{red}{\scriptsize (+3.6)}} 
& \makecell[c]{40.7 \\ \textcolor{BetterGreen}{\scriptsize (-135.9)}} 
\\
& Wan-R1 
& \makecell[c]{0.0 \\ \textcolor{gray}{\scriptsize (+0.0)}} 
& \makecell[c]{47.9 \\ \textcolor{BetterGreen}{\scriptsize (+47.9)}} 
& \makecell[c]{0.0 \\ \textcolor{gray}{\scriptsize (+0.0)}} 
& \makecell[c]{0.0 \\ \textcolor{gray}{\scriptsize (+0.0)}} 
& \makecell[c]{9.7 \\ \textcolor{BetterGreen}{\scriptsize (+2.8)}} 
& \makecell[c]{69.7 \\ \textcolor{BetterGreen}{\scriptsize (+57.2)}} 
& \makecell[c]{83.3 \\ \textcolor{BetterGreen}{\scriptsize (+51.4)}} 
& \makecell[c]{16.7 \\ \textcolor{BetterGreen}{\scriptsize (+5.6)}} 
& \makecell[c]{11.6 \\ \textcolor{BetterGreen}{\scriptsize (+5.0)}} 
& \makecell[c]{68.4 \\ \textcolor{BetterGreen}{\scriptsize (+59.3)}} 
& \makecell[c]{15.1 \\ \textcolor{BetterGreen}{\scriptsize (+2.3)}} 
& \makecell[c]{16.9 \\ \textcolor{BetterGreen}{\scriptsize (+7.7)}} 
& \makecell[c]{0.0 \\ \textcolor{BetterGreen}{\scriptsize (-388.7)}} 
& \makecell[c]{8.3 \\ \textcolor{BetterGreen}{\scriptsize (-57.8)}} 
& \makecell[c]{7.5 \\ \textcolor{red}{\scriptsize (+2.1)}} 
& \makecell[c]{9.3 \\ \textcolor{BetterGreen}{\scriptsize (-167.3)}} 
\\
\midrule

% ---------- 3D Maze ----------
\multirow{2}{*}{3D Maze} 
& Wan-SFT 
& \makecell[c]{0.0 \\ \textcolor{gray}{\scriptsize (+0.0)}} 
& \makecell[c]{5.6 \\ \textcolor{BetterGreen}{\scriptsize (+5.6)}} 
& \makecell[c]{65.3 \\ \textcolor{BetterGreen}{\scriptsize (+65.3)}} 
& \makecell[c]{0.0 \\ \textcolor{gray}{\scriptsize (+0.0)}} 
& \makecell[c]{38.9 \\ \textcolor{BetterGreen}{\scriptsize (+32.0)}} 
& \makecell[c]{31.9 \\ \textcolor{BetterGreen}{\scriptsize (+19.4)}} 
& \makecell[c]{100.0 \\ \textcolor{BetterGreen}{\scriptsize (+68.1)}} 
& \makecell[c]{20.8 \\ \textcolor{BetterGreen}{\scriptsize (+9.7)}} 
& \makecell[c]{6.8 \\ \textcolor{BetterGreen}{\scriptsize (+0.2)}} 
& \makecell[c]{20.6 \\ \textcolor{BetterGreen}{\scriptsize (+11.5)}} 
& \makecell[c]{93.5 \\ \textcolor{BetterGreen}{\scriptsize (+80.7)}} 
& \makecell[c]{15.0 \\ \textcolor{BetterGreen}{\scriptsize (+5.8)}} 
& \makecell[c]{108.2 \\ \textcolor{BetterGreen}{\scriptsize (-280.5)}} 
& \makecell[c]{10.9 \\ \textcolor{BetterGreen}{\scriptsize (-55.2)}} 
& \makecell[c]{3.9 \\ \textcolor{BetterGreen}{\scriptsize (-1.5)}} 
& \makecell[c]{80.6 \\ \textcolor{BetterGreen}{\scriptsize (-96.0)}} 
\\
& Wan-R1 
& \makecell[c]{0.0 \\ \textcolor{gray}{\scriptsize (+0.0)}} 
& \makecell[c]{6.9 \\ \textcolor{BetterGreen}{\scriptsize (+6.9)}} 
& \makecell[c]{94.4 \\ \textcolor{BetterGreen}{\scriptsize (+94.4)}} 
& \makecell[c]{0.0 \\ \textcolor{gray}{\scriptsize (+0.0)}} 
& \makecell[c]{41.7 \\ \textcolor{BetterGreen}{\scriptsize (+34.8)}} 
& \makecell[c]{39.2 \\ \textcolor{BetterGreen}{\scriptsize (+26.7)}} 
& \makecell[c]{100.0 \\ \textcolor{BetterGreen}{\scriptsize (+68.1)}} 
& \makecell[c]{23.6 \\ \textcolor{BetterGreen}{\scriptsize (+12.5)}} 
& \makecell[c]{5.4 \\ \textcolor{red}{\scriptsize (-1.2)}} 
& \makecell[c]{22.0 \\ \textcolor{BetterGreen}{\scriptsize (+12.9)}} 
& \makecell[c]{98.0 \\ \textcolor{BetterGreen}{\scriptsize (+85.2)}} 
& \makecell[c]{12.5 \\ \textcolor{BetterGreen}{\scriptsize (+3.3)}} 
& \makecell[c]{20.5 \\ \textcolor{BetterGreen}{\scriptsize (-368.2)}} 
& \makecell[c]{8.3 \\ \textcolor{BetterGreen}{\scriptsize (-57.8)}} 
& \makecell[c]{1.9 \\ \textcolor{BetterGreen}{\scriptsize (-3.5)}} 
& \makecell[c]{28.2 \\ \textcolor{BetterGreen}{\scriptsize (-148.4)}} 
\\
\midrule

% ---------- Sokoban ----------
\multirow{2}{*}{Sokoban} 
& Wan-SFT 
& \makecell[c]{0.0 \\ \textcolor{gray}{\scriptsize (+0.0)}} 
& \makecell[c]{1.4 \\ \textcolor{BetterGreen}{\scriptsize (+1.4)}} 
& \makecell[c]{0.0 \\ \textcolor{gray}{\scriptsize (+0.0)}} 
& \makecell[c]{4.2 \\ \textcolor{BetterGreen}{\scriptsize (+4.2)}} 
& \makecell[c]{0.5 \\ \textcolor{red}{\scriptsize (-6.4)}} 
& \makecell[c]{22.2 \\ \textcolor{BetterGreen}{\scriptsize (+9.7)}} 
& \makecell[c]{22.2 \\ \textcolor{red}{\scriptsize (-9.7)}} 
& \makecell[c]{69.4 \\ \textcolor{BetterGreen}{\scriptsize (+58.3)}} 
& \makecell[c]{15.7 \\ \textcolor{BetterGreen}{\scriptsize (+9.1)}} 
& \makecell[c]{23.7 \\ \textcolor{BetterGreen}{\scriptsize (+14.6)}} 
& \makecell[c]{15.7 \\ \textcolor{BetterGreen}{\scriptsize (+2.9)}} 
& \makecell[c]{44.3 \\ \textcolor{BetterGreen}{\scriptsize (+35.1)}} 
& \makecell[c]{46.3 \\ \textcolor{BetterGreen}{\scriptsize (-342.4)}} 
& \makecell[c]{34.4 \\ \textcolor{BetterGreen}{\scriptsize (-31.7)}} 
& \makecell[c]{20.1 \\ \textcolor{red}{\scriptsize (+14.7)}} 
& \makecell[c]{10.2 \\ \textcolor{BetterGreen}{\scriptsize (-166.4)}} 
\\
& Wan-R1 
& \makecell[c]{2.8 \\ \textcolor{BetterGreen}{\scriptsize (+2.8)}} 
& \makecell[c]{1.3 \\ \textcolor{BetterGreen}{\scriptsize (+1.3)}} 
& \makecell[c]{0.0 \\ \textcolor{gray}{\scriptsize (+0.0)}} 
& \makecell[c]{30.6 \\ \textcolor{BetterGreen}{\scriptsize (+30.6)}} 
& \makecell[c]{48.6 \\ \textcolor{BetterGreen}{\scriptsize (+41.7)}} 
& \makecell[c]{19.2 \\ \textcolor{BetterGreen}{\scriptsize (+6.7)}} 
& \makecell[c]{29.2 \\ \textcolor{red}{\scriptsize (-2.7)}} 
& \makecell[c]{68.1 \\ \textcolor{BetterGreen}{\scriptsize (+57.0)}} 
& \makecell[c]{21.1 \\ \textcolor{BetterGreen}{\scriptsize (+14.5)}} 
& \makecell[c]{26.1 \\ \textcolor{BetterGreen}{\scriptsize (+17.0)}} 
& \makecell[c]{15.4 \\ \textcolor{BetterGreen}{\scriptsize (+2.6)}} 
& \makecell[c]{46.9 \\ \textcolor{BetterGreen}{\scriptsize (+37.7)}} 
& \makecell[c]{6.1 \\ \textcolor{BetterGreen}{\scriptsize (-382.6)}} 
& \makecell[c]{29.2 \\ \textcolor{BetterGreen}{\scriptsize (-36.9)}} 
& \makecell[c]{8.6 \\ \textcolor{red}{\scriptsize (+3.2)}} 
& \makecell[c]{16.9 \\ \textcolor{BetterGreen}{\scriptsize (-159.7)}} 
\\
\midrule

% ---------- Trapfield ----------
\multirow{2}{*}{Trapfield} 
& Wan-SFT 
& \makecell[c]{0.0 \\ \textcolor{gray}{\scriptsize (+0.0)}} 
& \makecell[c]{0.0 \\ \textcolor{gray}{\scriptsize (+0.0)}} 
& \makecell[c]{0.0 \\ \textcolor{gray}{\scriptsize (+0.0)}} 
& \makecell[c]{0.0 \\ \textcolor{gray}{\scriptsize (+0.0)}} 
& \makecell[c]{93.1 \\ \textcolor{BetterGreen}{\scriptsize (+86.2)}} 
& \makecell[c]{40.3 \\ \textcolor{BetterGreen}{\scriptsize (+27.8)}} 
& \makecell[c]{79.2 \\ \textcolor{BetterGreen}{\scriptsize (+47.3)}} 
& \makecell[c]{6.9 \\ \textcolor{red}{\scriptsize (-4.2)}} 
& \makecell[c]{10.9 \\ \textcolor{BetterGreen}{\scriptsize (+4.3)}} 
& \makecell[c]{12.9 \\ \textcolor{BetterGreen}{\scriptsize (+3.8)}} 
& \makecell[c]{14.7 \\ \textcolor{BetterGreen}{\scriptsize (+1.9)}} 
& \makecell[c]{10.0 \\ \textcolor{BetterGreen}{\scriptsize (+0.8)}} 
& \makecell[c]{57.5 \\ \textcolor{BetterGreen}{\scriptsize (-331.2)}} 
& \makecell[c]{16.8 \\ \textcolor{BetterGreen}{\scriptsize (-49.3)}} 
& \makecell[c]{11.4 \\ \textcolor{red}{\scriptsize (+6.0)}} 
& \makecell[c]{57.8 \\ \textcolor{BetterGreen}{\scriptsize (-118.8)}} 
\\

& Wan-R1 
& \makecell[c]{0.0 \\ \textcolor{gray}{\scriptsize (+0.0)}} 
& \makecell[c]{0.0 \\ \textcolor{gray}{\scriptsize (+0.0)}} 
& \makecell[c]{0.0 \\ \textcolor{gray}{\scriptsize (+0.0)}} 
& \makecell[c]{0.0 \\ \textcolor{gray}{\scriptsize (+0.0)}} 
& \makecell[c]{93.1 \\ \textcolor{BetterGreen}{\scriptsize (+86.2)}} 
& \makecell[c]{34.7 \\ \textcolor{BetterGreen}{\scriptsize (+22.2)}} 
& \makecell[c]{73.6 \\ \textcolor{BetterGreen}{\scriptsize (+41.7)}} 
& \makecell[c]{6.9 \\ \textcolor{red}{\scriptsize (-4.2)}} 
& \makecell[c]{10.8 \\ \textcolor{BetterGreen}{\scriptsize (+4.2)}} 
& \makecell[c]{12.8 \\ \textcolor{BetterGreen}{\scriptsize (+3.7)}} 
& \makecell[c]{14.4 \\ \textcolor{BetterGreen}{\scriptsize (+1.6)}} 
& \makecell[c]{9.3 \\ \textcolor{BetterGreen}{\scriptsize (+0.1)}} 
& \makecell[c]{10.8 \\ \textcolor{BetterGreen}{\scriptsize (-377.9)}} 
& \makecell[c]{14.7 \\ \textcolor{BetterGreen}{\scriptsize (-51.4)}} 
& \makecell[c]{4.4 \\ \textcolor{BetterGreen}{\scriptsize (-1.0)}} 
& \makecell[c]{24.0 \\ \textcolor{BetterGreen}{\scriptsize (-152.6)}} 
\\

\bottomrule
\end{tabular}
}
\caption{
Performance comparison between the baseline model (Wan2.2-TI2V-5B), task-specific Wan-SFT models, and our method across four game types: Base, Irreg, 3D, and Soko. Each cell displays the absolute performance alongside the relative change from the baseline across four metrics (EM, SR, PR, SD).
}
\label{tab:type_gen}
\end{table*}

Table~\ref{tab:type_gen} presents cross-task generalization results when training on a single maze type. GRPO consistently improves transfer to unseen maze types compared to SFT, suggesting that RL encourages learning of more abstract spatial reasoning skills that generalize beyond the specific task structure encountered during training.

\subsection{Texture Generalization.}

\begin{table*}[h]
\centering
\setlength{\tabcolsep}{3pt}
\renewcommand{\arraystretch}{0.75}
\resizebox{0.75\linewidth}{!}{
\footnotesize
\begin{tabular}{l|l|ccc|ccc|ccc|ccc}
\toprule
\textbf{Task} & 
& \multicolumn{3}{c|}{\textbf{EM}} 
& \multicolumn{3}{c|}{\textbf{SR}} 
& \multicolumn{3}{c|}{\textbf{PR}} 
& \multicolumn{3}{c}{\textbf{SD}} \\
& & Raw & Sk2 & Sk3 & Raw & Sk2 & Sk3 & Raw & Sk2 & Sk3 & Raw & Sk2 & Sk3 \\
\midrule

% ---- Base ----
\multirow{4}{*}{Base}
& Base   & 0.0 & 0.0 & 0.0 & 6.9 & 4.2 & 2.8 & 6.6 & 4.6 & 9.4 & 388.7 & 11.7 & 14.9 \\
& \cellcolor{cyan!15}SFT    & \cellcolor{cyan!15}33.3 & \cellcolor{cyan!15}1.4 & \cellcolor{cyan!15}23.6 & \cellcolor{cyan!15}76.4 & \cellcolor{cyan!15}38.9 & \cellcolor{cyan!15}68.1 & \cellcolor{cyan!15}60.6 & \cellcolor{cyan!15}12.3 & \cellcolor{cyan!15}51.0 & \cellcolor{cyan!15}10.3 & \cellcolor{cyan!15}19.0 & \cellcolor{cyan!15}9.0 \\
& \cellcolor{yellow!20}R1     & \cellcolor{yellow!20}61.1 & \cellcolor{yellow!20}31.9 & \cellcolor{yellow!20}31.9 & \cellcolor{yellow!20}75.0 & \cellcolor{yellow!20}70.8 & \cellcolor{yellow!20}70.8 & \cellcolor{yellow!20}78.6 & \cellcolor{yellow!20}52.5 & \cellcolor{yellow!20}52.5 & \cellcolor{yellow!20}5.2 & \cellcolor{yellow!20}27.1 & \cellcolor{yellow!20}27.1 \\
& $\Delta$ & {\scriptsize\textcolor{BetterGreen}{+61.1}} & {\scriptsize\textcolor{BetterGreen}{+31.9}} & {\scriptsize\textcolor{BetterGreen}{+31.9}} & {\scriptsize\textcolor{BetterGreen}{+68.1}} & {\scriptsize\textcolor{BetterGreen}{+66.6}} & {\scriptsize\textcolor{BetterGreen}{+68.0}} & {\scriptsize\textcolor{BetterGreen}{+72.0}} & {\scriptsize\textcolor{BetterGreen}{+47.9}} & {\scriptsize\textcolor{BetterGreen}{+43.1}} & {\scriptsize\textcolor{BetterGreen}{-383.5}} & {\scriptsize\textcolor{BetterGreen}{+15.4}} & {\scriptsize\textcolor{BetterGreen}{+12.2}} \\
\midrule

% ---- Irreg ----
\multirow{4}{*}{Irreg}
& Base   & 0.0 & 0.0 & 0.0 & 12.5 & 4.2 & 2.1 & 9.1 & 12.0 & 47.8 & 66.1 & 39.5 & 42.0 \\
& \cellcolor{cyan!15}SFT    & \cellcolor{cyan!15}56.9 & \cellcolor{cyan!15}22.2 & \cellcolor{cyan!15}15.3 & \cellcolor{cyan!15}69.4 & \cellcolor{cyan!15}15.3 & \cellcolor{cyan!15}23.6 & \cellcolor{cyan!15}71.6 & \cellcolor{cyan!15}36.5 & \cellcolor{cyan!15}26.2 & \cellcolor{cyan!15}2.4 & \cellcolor{cyan!15}5.1 & \cellcolor{cyan!15}7.7 \\
& \cellcolor{yellow!20}R1     & \cellcolor{yellow!20}47.9 & \cellcolor{yellow!20}12.3 & \cellcolor{yellow!20}12.3 & \cellcolor{yellow!20}69.7 & \cellcolor{yellow!20}13.6 & \cellcolor{yellow!20}13.6 & \cellcolor{yellow!20}68.4 & \cellcolor{yellow!20}19.3 & \cellcolor{yellow!20}19.3 & \cellcolor{yellow!20}8.3 & \cellcolor{yellow!20}4.2 & \cellcolor{yellow!20}4.2 \\
& $\Delta$ & {\scriptsize\textcolor{BetterGreen}{+47.9}} & {\scriptsize\textcolor{BetterGreen}{+12.3}} & {\scriptsize\textcolor{BetterGreen}{+12.3}} & {\scriptsize\textcolor{BetterGreen}{+57.2}} & {\scriptsize\textcolor{BetterGreen}{+9.4}} & {\scriptsize\textcolor{BetterGreen}{+11.5}} & {\scriptsize\textcolor{BetterGreen}{+59.3}} & {\scriptsize\textcolor{BetterGreen}{+7.3}} & {\scriptsize\textcolor{red}{-28.5}} & {\scriptsize\textcolor{BetterGreen}{-57.8}} & {\scriptsize\textcolor{BetterGreen}{-35.3}} & {\scriptsize\textcolor{BetterGreen}{-37.8}} \\
\midrule

% ---- 3D ----
\multirow{4}{*}{3D}
& Base   & 0.0 & 0.0 & 0.0 & 0.0 & 4.2 & 4.2 & 7.1 & 8.9 & 8.3 & -- & 73.6 & 59.3 \\
& \cellcolor{cyan!15}SFT    & \cellcolor{cyan!15}65.3 & \cellcolor{cyan!15}43.1 & \cellcolor{cyan!15}50.0 & \cellcolor{cyan!15}100.0 & \cellcolor{cyan!15}97.2 & \cellcolor{cyan!15}100.0 & \cellcolor{cyan!15}93.5 & \cellcolor{cyan!15}83.8 & \cellcolor{cyan!15}86.8 & \cellcolor{cyan!15}3.9 & \cellcolor{cyan!15}6.4 & \cellcolor{cyan!15}5.9 \\
& \cellcolor{yellow!20}R1     & \cellcolor{yellow!20}94.4 & \cellcolor{yellow!20}79.2 & \cellcolor{yellow!20}79.2 & \cellcolor{yellow!20}100.0 & \cellcolor{yellow!20}100.0 & \cellcolor{yellow!20}100.0 & \cellcolor{yellow!20}98.0 & \cellcolor{yellow!20}92.2 & \cellcolor{yellow!20}92.2 & \cellcolor{yellow!20}1.9 & \cellcolor{yellow!20}2.9 & \cellcolor{yellow!20}2.9 \\
& $\Delta$ & {\scriptsize\textcolor{BetterGreen}{+94.4}} & {\scriptsize\textcolor{BetterGreen}{+79.2}} & {\scriptsize\textcolor{BetterGreen}{+79.2}} & {\scriptsize\textcolor{BetterGreen}{+100.0}} & {\scriptsize\textcolor{BetterGreen}{+95.8}} & {\scriptsize\textcolor{BetterGreen}{+95.8}} & {\scriptsize\textcolor{BetterGreen}{+90.9}} & {\scriptsize\textcolor{BetterGreen}{+83.3}} & {\scriptsize\textcolor{BetterGreen}{+83.9}} & {\scriptsize\textcolor{gray}{--}} & {\scriptsize\textcolor{BetterGreen}{-70.7}} & {\scriptsize\textcolor{BetterGreen}{-56.4}} \\
\midrule

% ---- Soko ----
\multirow{4}{*}{Soko}
& Base   & 0.0 & 0.0 & 0.0 & 31.9 & 2.8 & 7.1 & 12.8 & 9.3 & 8.1 & 5.4 & -- & -- \\
& \cellcolor{cyan!15}SFT    & \cellcolor{cyan!15}4.2 & \cellcolor{cyan!15}0.0 & \cellcolor{cyan!15}1.4 & \cellcolor{cyan!15}69.4 & \cellcolor{cyan!15}34.7 & \cellcolor{cyan!15}58.3 & \cellcolor{cyan!15}44.3 & \cellcolor{cyan!15}21.0 & \cellcolor{cyan!15}14.1 & \cellcolor{cyan!15}10.2 & \cellcolor{cyan!15}58.6 & \cellcolor{cyan!15}82.6 \\
& \cellcolor{yellow!20}R1     & \cellcolor{yellow!20}30.6 & \cellcolor{yellow!20}0.0 & \cellcolor{yellow!20}0.0 & \cellcolor{yellow!20}68.1 & \cellcolor{yellow!20}70.8 & \cellcolor{yellow!20}70.8 & \cellcolor{yellow!20}46.9 & \cellcolor{yellow!20}14.7 & \cellcolor{yellow!20}14.7 & \cellcolor{yellow!20}16.9 & \cellcolor{yellow!20}49.1 & \cellcolor{yellow!20}49.1 \\
& $\Delta$ & {\scriptsize\textcolor{BetterGreen}{+30.6}} & {\scriptsize\textcolor{gray}{+0.0}} & {\scriptsize\textcolor{gray}{+0.0}} & {\scriptsize\textcolor{BetterGreen}{+36.2}} & {\scriptsize\textcolor{BetterGreen}{+68.0}} & {\scriptsize\textcolor{BetterGreen}{+63.7}} & {\scriptsize\textcolor{BetterGreen}{+34.1}} & {\scriptsize\textcolor{BetterGreen}{+5.4}} & {\scriptsize\textcolor{BetterGreen}{+6.6}} & {\scriptsize\textcolor{red}{+11.5}} & {\scriptsize\textcolor{gray}{--}} & {\scriptsize\textcolor{gray}{--}} \\
\midrule

% ---- Trap ----
\multirow{4}{*}{Trap}
& Base   & 0.0 & 0.0 & 0.0 & 11.1 & 0.0 & 0.0 & 9.2 & 9.0 & 7.5 & 176.6 & -- & -- \\
& \cellcolor{cyan!15}SFT    & \cellcolor{cyan!15}38.9 & \cellcolor{cyan!15}9.7 & \cellcolor{cyan!15}0.0 & \cellcolor{cyan!15}100.0 & \cellcolor{cyan!15}38.9 & \cellcolor{cyan!15}1.4 & \cellcolor{cyan!15}79.1 & \cellcolor{cyan!15}29.0 & \cellcolor{cyan!15}9.9 & \cellcolor{cyan!15}3.9 & \cellcolor{cyan!15}8.7 & \cellcolor{cyan!15}18.5 \\
& \cellcolor{yellow!20}R1     & \cellcolor{yellow!20}90.3 & \cellcolor{yellow!20}0.0 & \cellcolor{yellow!20}0.0 & \cellcolor{yellow!20}100.0 & \cellcolor{yellow!20}1.4 & \cellcolor{yellow!20}1.4 & \cellcolor{yellow!20}97.7 & \cellcolor{yellow!20}8.5 & \cellcolor{yellow!20}8.5 & \cellcolor{yellow!20}2.4 & \cellcolor{yellow!20}0.0 & \cellcolor{yellow!20}0.0 \\
& $\Delta$ & {\scriptsize\textcolor{BetterGreen}{+90.3}} & {\scriptsize\textcolor{gray}{+0.0}} & {\scriptsize\textcolor{gray}{+0.0}} & {\scriptsize\textcolor{BetterGreen}{+88.9}} & {\scriptsize\textcolor{BetterGreen}{+1.4}} & {\scriptsize\textcolor{BetterGreen}{+1.4}} & {\scriptsize\textcolor{BetterGreen}{+88.5}} & {\scriptsize\textcolor{red}{-0.5}} & {\scriptsize\textcolor{BetterGreen}{+1.0}} & {\scriptsize\textcolor{BetterGreen}{-174.2}} & {\scriptsize\textcolor{gray}{--}} & {\scriptsize\textcolor{gray}{--}} \\

\bottomrule
\end{tabular}}

\caption{
Texture generalization performance across different visual skins.
For each task, we report the baseline (Base), \colorbox{cyan!15}{Wan-SFT}, and \colorbox{yellow!20}{Wan-R1} across three texture conditions (Raw, Sk2=Skin2, Sk3=Skin3), with $\Delta$ indicating Wan-R1 improvement over baseline. See texture sample in~\autoref{fig:vrbench_sample}.
}
\vspace{-1.5em}
\label{tab:texture_gen}
\end{table*}
Table~\ref{tab:texture_gen} evaluates performance on unseen visual textures. The RL-tuned model demonstrates stronger texture invariance, particularly on styles that differ substantially from training. This indicates that RL encourages the model to focus on structural maze features rather than superficial visual patterns.

\subsection{Video Quality Preservation}

\begin{table}[t]
\centering
\setlength{\tabcolsep}{0.9mm}{
\resizebox{0.6\textwidth}{!}{
\begin{tabular}{c|l|ccccc}
\toprule
% ----- Top header row -----
\multicolumn{2}{c|}{\multirow{2}{*}{\textbf{Method}}} &
\multicolumn{5}{c}{\textbf{MF}~($\uparrow$)} \\
% ----- Second header row -----
& & Base & Irreg & Trap & 3D & Soko \\ \midrule
\multirow{11}{*}{\rotatebox{90}{\textbf{General Video Model}}} & \multicolumn{1}{c}{\textbf{Closed-Source}} \\
& Veo-3.1-fast          & 43.2 & 86.3 & 22.5 & 69.1 & 63.5 \\ 
& Veo-3.1-pro           & 80.5 & 89.3 & 82.2 & 73.4 & \textbf{95.8} \\
& Sora-2                & \textbf{96.5} & 97.2 & \textbf{97.2} & \underline{95.4} & \underline{95.2} \\ 
% & Sora-2-pro            &  &  &  &  &  \\ 
& kling-v1              & 55.5 & 73.8 & 54.4 & 84.9 & 72.9 \\
& Seedance-1.0-pro      & 87.7 & 97.2 & 64.3 & 86.3 & 83.0 \\
& MiniMax-Hailuo-2.3    & \underline{92.4} & 93.4 & 91.4 & 94.9 & 93.3 \\ \cline{2-7} \noalign{\vskip 3pt}
& \multicolumn{1}{c}{\textbf{Open-Source}} & \\
& Wan2.5-i2v-preview    & 69.2 & 74.8 & 70.6 & 82.7 & 90.4 \\
& Wan2.2-TI2V-5B$^\Diamond$ & 85.5 &\underline{97.4} & 83.7 & 94.7 & 93.5 \\
& Wan-SFT~\cite{yang2025reasoningvideoevaluationvideo}                       & 91.2 & 98.1 & \underline{93.3} & \textbf{95.7} & 94.1 \\ \midrule 
\multirow{1}{*}{\rotatebox{90}{\textbf{Train}}} 
& \textbf{Wan-R1}                        & 94.1 & 98.1 & 94.9 & 95.6 & 73.0 \\
& \multicolumn{1}{l|}{\textbf{-Wan2.2-TI2V-5B}$^\Diamond$} & \cellcolor{yellow!20}{+8.6} & \cellcolor{yellow!20}{+0.7} & \cellcolor{yellow!20}{+11.2} & \cellcolor{yellow!20}{+0.9} & \cellcolor{yellow!20}{-20.5} \\
\bottomrule
\end{tabular}}}
\caption{MF denotes Maze Fidelity.}
\vspace{-1em}
\label{tab:quality}
\end{table}

Table~\ref{tab:quality} confirms that RL training does not degrade video quality in most cases. Maze Fidelity improves from 91.2\% to 94.1\% on Regular Maze and from 93.3\% to 94.9\% on Trapfield, while remaining stable on Irregular and 3D Maze. The decrease on Sokoban (94.1\% to 73.0\%) warrants further investigation but does not reflect a systematic quality degradation. These results indicate that performance improvements come from better reasoning rather than exploiting quality shortcuts or generating visually deceptive outputs.

\subsection{Cold Start}

Applying Flow-GRPO directly to the base model (Wan2.2-TI2V-5B) without prior supervised fine-tuning yields poor results, as the base model achieves near-zero exact match across all maze types. The results are presented in~\autoref{tab:cold_start}. This highlights a standard characteristic of reinforcement learning: the policy requires a foundational task understanding before RL can effectively guide exploration~\citep{yue2025doesreinforcementlearningreally,feng2025videor1reinforcingvideoreasoning}. Because our verifiable rewards enforce strict logical correctness, the reward is effectively too sparse near a cold start because the base policy rarely reaches informative regions. Therefore, SFT acts as a necessary bootstrap, placing the model within a region of the reward landscape where policy gradient updates become meaningful. Future work could bridge this gap through curriculum learning on simplified environments or offline preference optimization.

\begin{table}[h]
% %\vspace{-2em}
\centering
\small
\begin{subtable}[h]{0.48\textwidth}
\centering
\begin{tabular}{l|cccc}
\toprule
     & EM & SR & PR & SD \\
\midrule
w.o SFT    & 0.0 & 2.8 & 4.0 & 41.1 \\
w. SFT & 61.1 & 75.0 & 78.6 & 5.2
\\
\bottomrule
\end{tabular}
\caption{Effect of Cold Start.}
\label{tab:cold_start}
\end{subtable}%
\hfill
\begin{subtable}[h]{0.48\textwidth}
\centering
\begin{tabular}{l|cccc}
\toprule
$S_{\text{train}}$ & EM & SR & PR & SD \\
\midrule
Qwen2.5-VL-7B-Instruct & 11.1 & 55.6 & 42.4 & 2.2 \\
EM & 58.3 & 75.0 & 79.3 & 4.4 \\
PR \& EM & 61.1 & 75.0 & 78.6 & 5.2 \\
\bottomrule
\end{tabular}
\caption{Effect of reward models.}
\label{tab:reward_models}
\end{subtable}
\caption{Ablation studies on cold start (left) and reward models (right).}
%\vspace{-2em}
\end{table}

\subsection{Test-Time Scaling}

\begin{figure}[h]
\centering
\includegraphics[width=0.65\columnwidth]{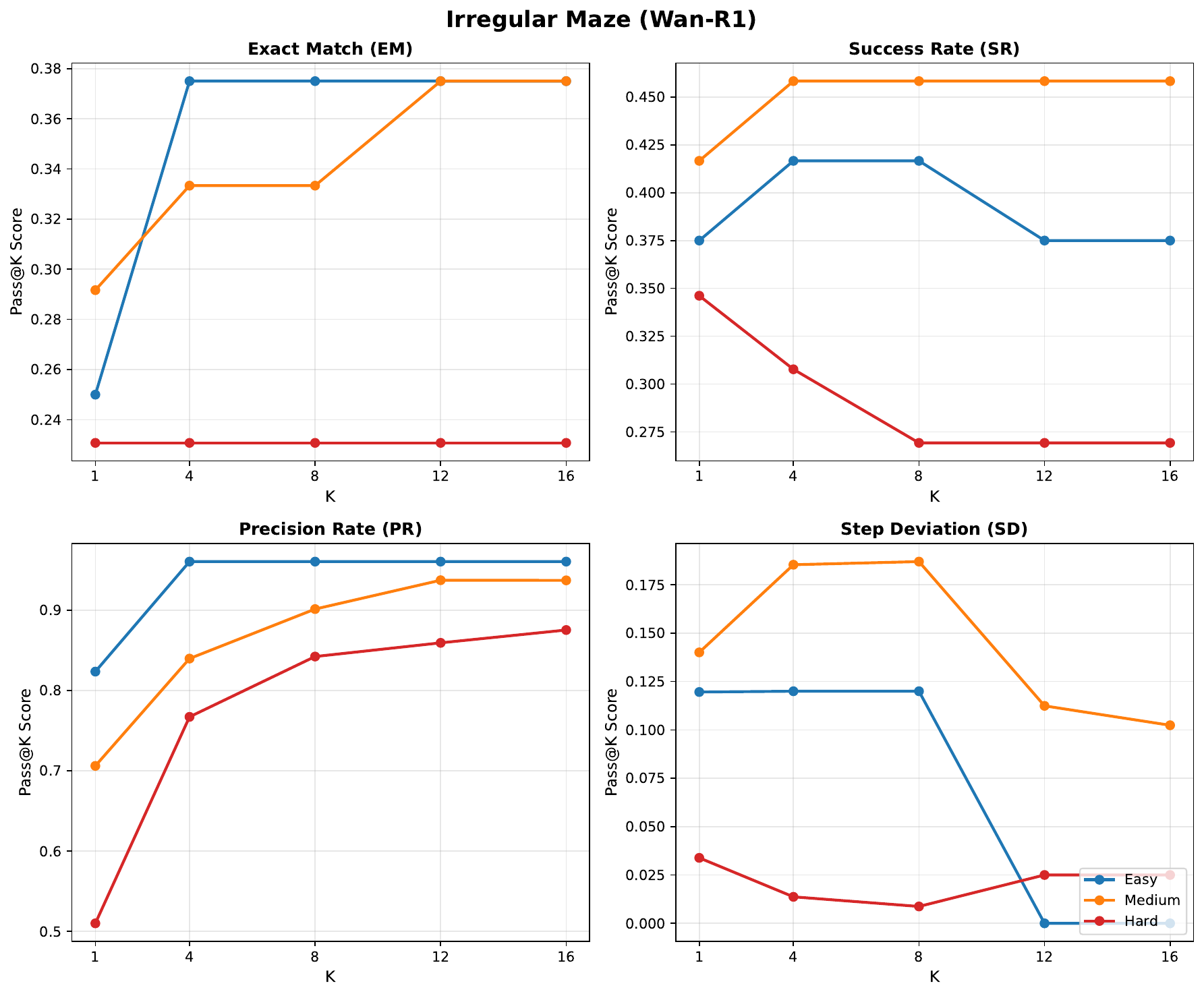} 
% %\vspace{-2em}
\caption{Test-time scaling performance of Wan-R1 on Irregular Maze tasks. Performance is evaluated across varying numbers of samples ($K \in \{1, 4, 8, 12, 16\}$) and three difficulty levels (Easy, Medium, Hard). RL improves base capability (K=1) while maintaining beneficial scaling with increased sampling.}
\label{fig:irregular-scaling}
%\vspace{-1.5em}
\end{figure}

Figure~\ref{fig:irregular-scaling} examines whether RL genuinely improves model capability or merely benefits from best-of-K sampling. We evaluate Wan-R1 with varying numbers of samples $K \in \{1, 4, 8, 12, 16\}$ across difficulty levels. The results reveal that performance continues to improve with larger K, demonstrating that the model maintains meaningful diversity in its generations. Importantly, even at K=1 (no sampling advantage), Wan-R1 substantially outperforms the SFT baseline, confirming that RL training improves the underlying policy rather than simply increasing variance for lucky samples. The scaling curves show diminishing returns beyond K=8, suggesting this as a practical inference budget for deployment.

\subsection{Reward Design Analysis}
\label{sec:reward_analysis}

We conduct a systematic study of reward design choices. This analysis serves two purposes: validating our 
multi-component verifiable reward and providing general insights on reward design in video reasoning. The results are presented in~\autoref{tab:reward_models}.

\begin{figure}[h]
\centering
\includegraphics[width=0.75\columnwidth]{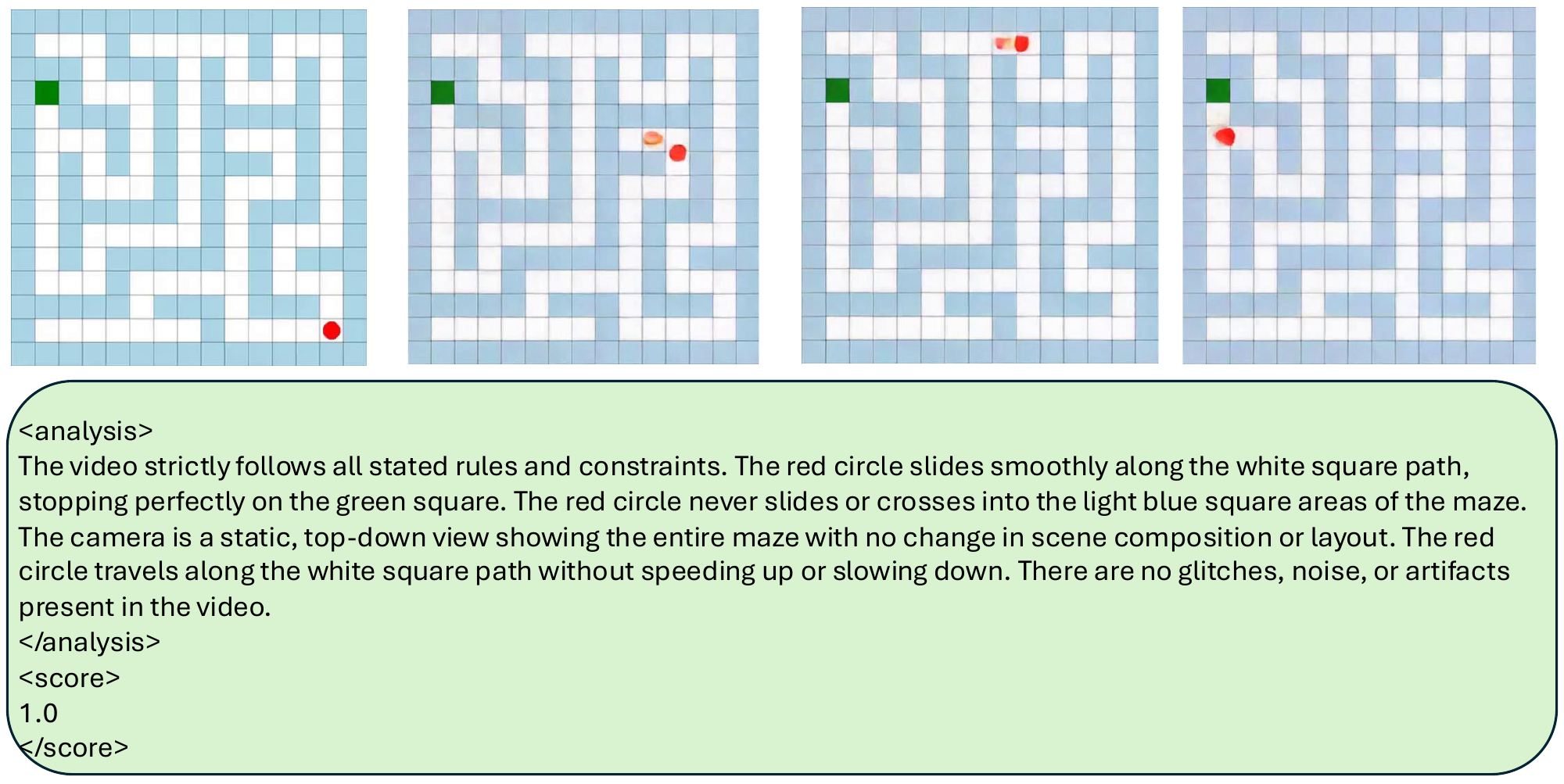}
\caption{Given the video output along with the evaluation prompt describing 
maze-solving rules, Qwen2.5-VL-7B-Instruct produces a step-by-step analysis 
and assigns a reward score. Although the generated video exhibits visible 
glitches and degraded visual quality, the model assigns a perfect score of 
1.0. This illustrates the core vulnerability of VLM-based reward models: 
rather than detecting actual visual artifacts, the model hallucinates quality 
based on high-level semantic cues---a visible agent, a recognizable maze 
structure, and apparent goal-directed motion---and incorrectly concludes that 
the video contains ``no glitches, noise, or artifacts.''}
\label{fig:qwen_judge}
\end{figure}

\paragraph{When reward models are not reliable}
We use Qwen2.5-VL-7B-Instruct as a reward model, prompting it to assess whether the generated video correctly solves the maze.~\autoref{fig:qwen_judge} reveals a fundamental failure mode: despite the generated video exhibiting visible glitches and degraded visual quality, the model produces a confident analysis stating that the agent moves smoothly with ``no glitches, noise, or artifacts present,'' and assigns a perfect reward score of 1.0. Rather than detecting actual visual fidelity, the model appears to hallucinate quality based on the presence of high-level semantic cues — a visible agent, a recognizable maze structure, and apparent goal-directed motion — that superficially resemble a correct solution. This confirms that VLM-based reward models cannot reliably assess the fine-grained visual properties of generated videos, as the high-dimensional, continuous nature of video outputs makes such models particularly vulnerable to exploitation. Unlike text-based RL where outputs are discrete and verifiable, video quality artifacts are subtle enough to evade VLM scrutiny entirely, rendering learned reward models an unreliable training signal.

\paragraph{Sparse rewards provide insufficient learning signal.}
Using only binary success (exact match ($R_\text{EM}$) as 
reward yields limited improvement over SFT. With $R_\text{EM}$ alone, the reward is sparse---few 
generated trajectories exactly match the optimal path, so most samples 
receive zero reward, providing negligible gradient signal for early 
training.

\paragraph{Dense verifiable rewards enable stable learning.}
The precision reward $R_\text{PR}$ provides the critical dense signal: 
it rewards partial progress along the correct path, creating a smooth 
reward landscape that guides optimization. Training with $R_\text{PR} + R_\textbf{PR}$ 
 achieves better EM, outperforming sparse 
alternatives. Adding $R_\text{EM}$ provides a bonus for complete 
solutions, encouraging the model to ``close out'' trajectories rather 
than settling for partial correctness.

\section{VR Bench}
\label{app:vr_bench_details}

\begin{figure}[h]
\centering
\includegraphics[width=0.85\columnwidth]{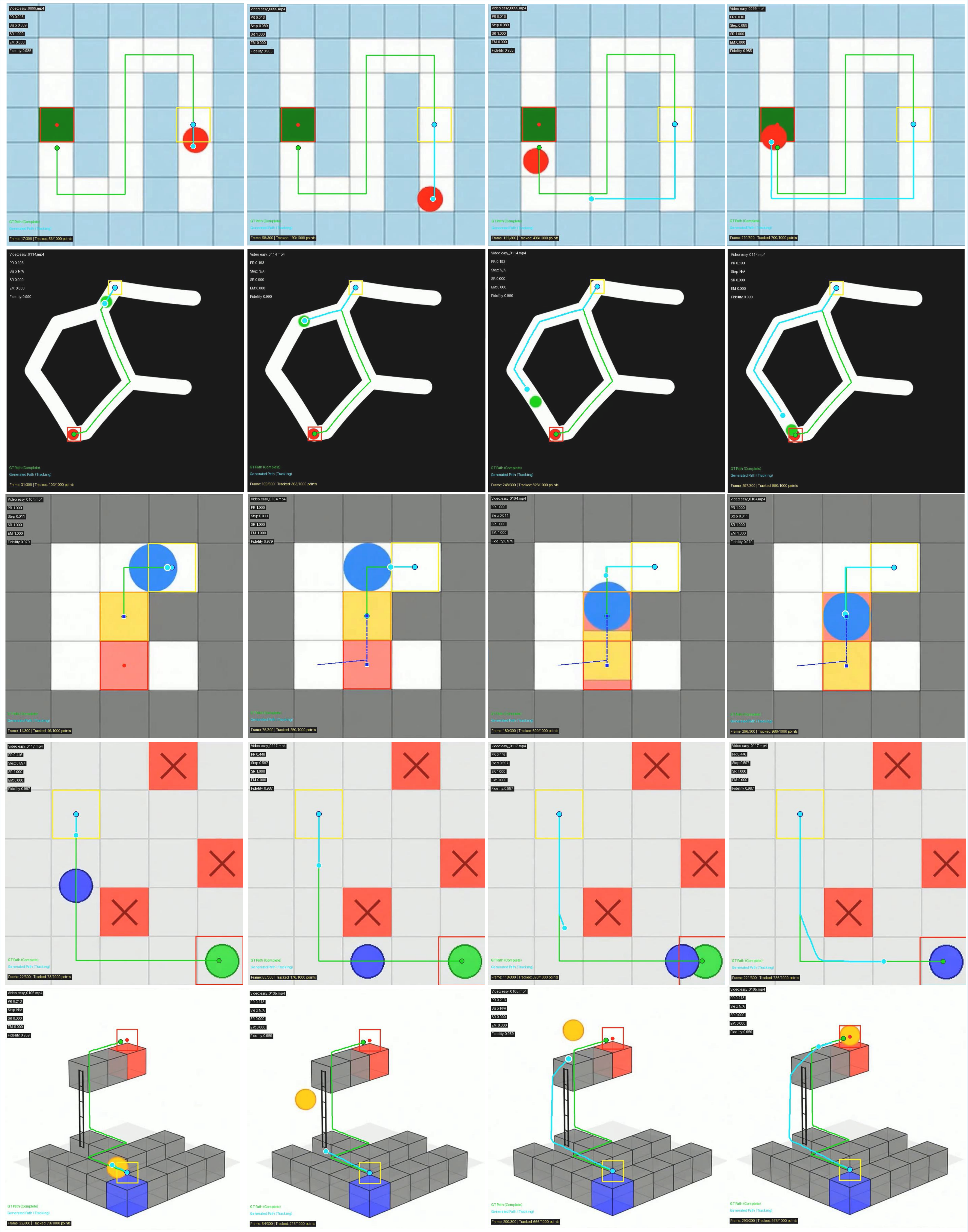} 
\caption{Trajectory tracking across five maze types (This figure is from~\citep{yang2025reasoningvideoevaluationvideo}). Green: ground-truth trajectory; Blue: tracked trajectory from generated video. Columns show temporally ordered frames illustrating agent motion over time.}
\label{fig:reasoning_traj}
\end{figure}

\begin{figure}[h]
\centering
\includegraphics[width=1\columnwidth]{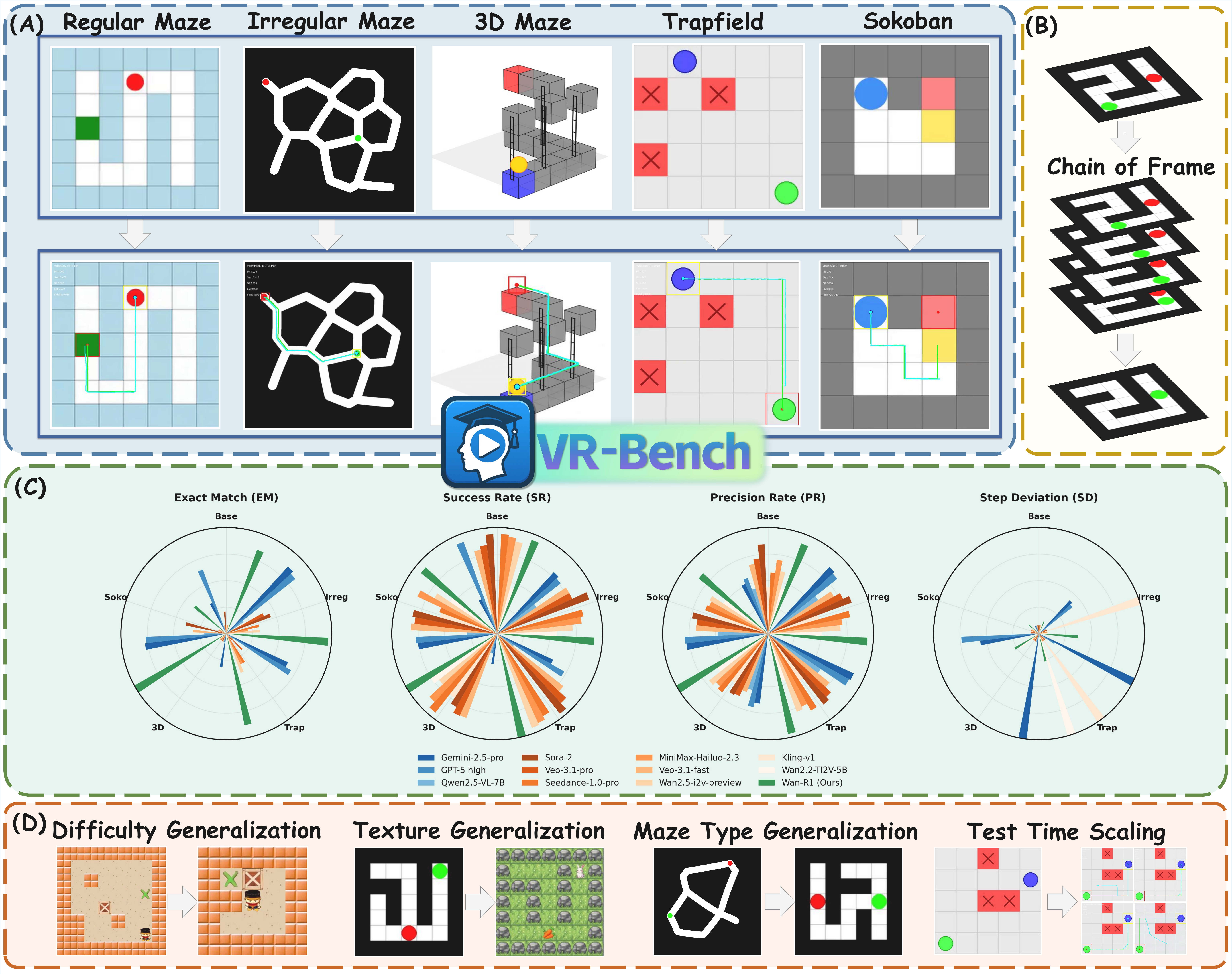} 
% %\vspace{-2em}
\captionof{figure}{Overview of VR-Bench (This figure is from VR-Bench~\citep{yang2025reasoningvideoevaluationvideo})
(A) Maze Types: Five maze types—Regular, Irregular, 3D, Trapfield, and Sokoban—spanning 2D/3D settings and diverse task structures.
(B) Reasoning Paradigm: Chain-of-frame reasoning \citep{wiedemer2025video} requiring frame-by-frame inferences for sequential visual reasoning.
(C) Benchmark Performance: Leading VLMs evaluated on four metrics across all maze types.
(D) Additional Analysis: Evaluations on difficulty, texture, and maze-type generalization, plus test-time scaling.}
\label{fig:vrbench_sample}

\end{figure}

In this section, we show the trajectory samples and game samples from VR-Bench in~\autoref{fig:reasoning_traj} and~\autoref{fig:vrbench_sample} respectively.

We conduct experiments on VR-Bench~\citep{yang2025reasoningvideoevaluationvideo}, a benchmark designed to evaluate video models' reasoning capabilities through maze-solving tasks. As illustrated in Figure~\ref{fig:vrbench_sample}, VR-Bench comprises five maze types of increasing complexity: Regular Maze (grid-based pathfinding), Irregular Maze (curved paths), 3D Maze (depth perception), Trapfield (obstacle avoidance), and Sokoban (box-pushing puzzles). Each maze type includes three difficulty levels and multiple visual textures, enabling systematic evaluation of generalization across task complexity, visual appearance, and reasoning requirements.

\section{Target Bench}
\label{app:target_bench_details}

\begin{figure}[h]
\centering
\includegraphics[width=0.85\columnwidth]{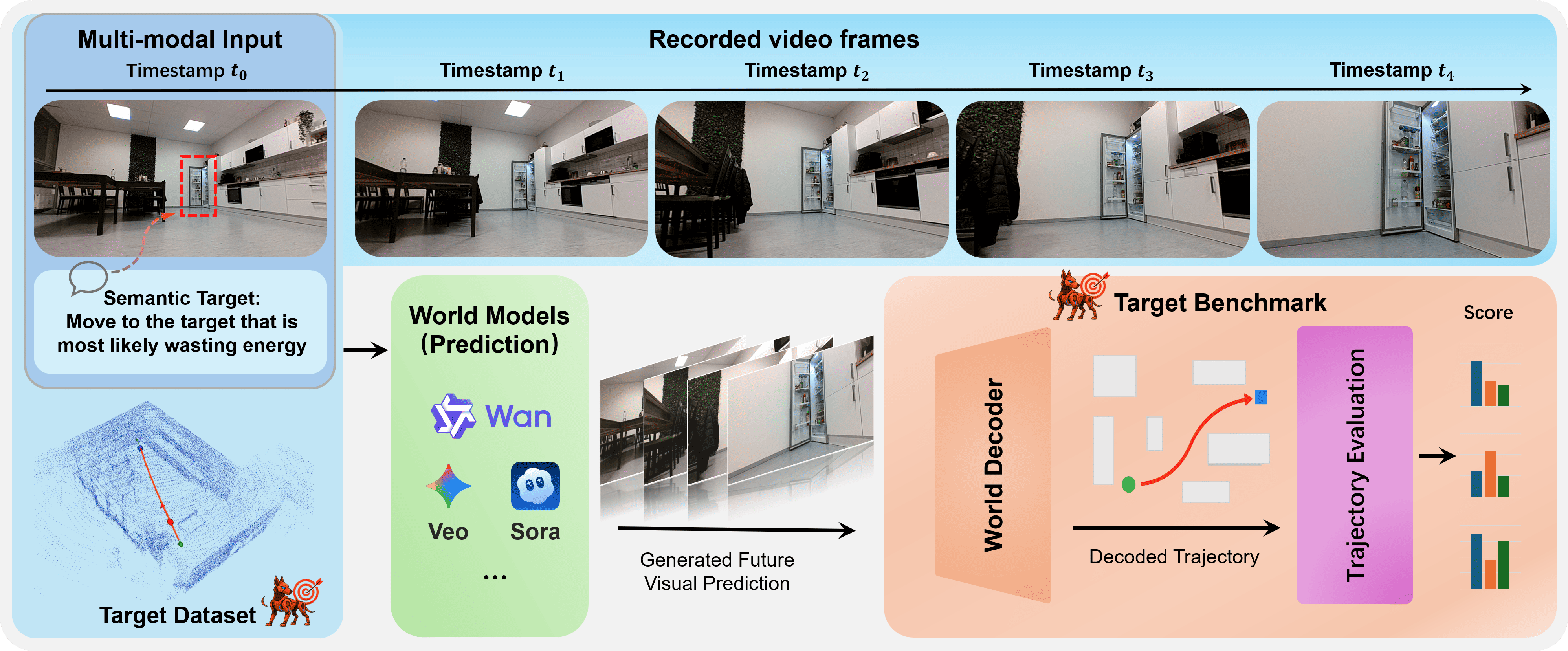}
\captionof{figure}{Overview of Target-Bench (This figure is from Target-Bench~\citep{wang2025targetbench}). Target-Bench introduces a quadruped-robot dataset and benchmark for assessing world models in mapless navigation toward text-specified goals with implicit semantic content. Given a single camera frame and a natural language description of the desired target state, world models must synthesize a future video depicting the trajectory to that goal. A world decoder then recovers the planned path from the generated video, which is evaluated against ground-truth maneuvers collected from a human-operated quadruped.}
\label{fig:reasoning_traj_target}
\end{figure}

In this section, we show the overview pipeline of Target-Bench in~\autoref{fig:reasoning_traj_target}.

% \section{LLM Usage Statement}
% \label{app:llm_usage}
% We acknowledge the use of large language models in the preparation of this work. Specifically, LLMs were employed to assist with experimental implementation and code debugging, as well as to improve grammar and clarity throughout the manuscript.

% \clearpage
% \section{Prompt Template}
% \label{app:prompts}
% All prompt we used in experiments are from~\citep{yang2025reasoningvideoevaluationvideo} as shown below. Each game has different prompt.

% \input{Prompt_Template/videoprompt1}

% \input{Prompt_Template/videoprompt2}

% \input{Prompt_Template/videoprompt3}

% \input{Prompt_Template/vlmprompt1}

% \input{Prompt_Template/vlmprompt2}

% \input{Prompt_Template/vlmprompt3}
 
% ---- Bibliography ----
%
% BibTeX users should specify bibliography style 'splncs04'.
% References will then be sorted and formatted in the correct style.
%
\bibliographystyle{splncs04}
\bibliography{main}
\end{document}